\documentclass[]{article}
\usepackage{style/proceed2e}

\usepackage{amssymb, amsmath, amsthm}
\usepackage{graphicx}
\usepackage{color}
\usepackage{xspace}
\usepackage[tight]{subfigure}
\usepackage[ruled,vlined]{algorithm2e}
\usepackage{natbib}
\usepackage{times}
\usepackage{url}

\newcommand{\lessdenselist}{
  \itemsep -3pt 
}

\newcommand{\denselistbib}{
  \itemsep 0pt\topsep-4pt\partopsep-5pt
}
\newcommand{\tightsection}[1]{\vspace{-3mm}
\section{\uppercase{#1}}\vspace{-2mm}}

\newcommand{\tightsubsection}[1]{
\vspace{-2mm}
\subsection{\uppercase{#1}}
\vspace{-2mm}
}

\graphicspath{{figures/}}

\numberwithin{equation}{section}

\newtheorem{dfn}{Definition}[section]

\newtheorem{thm}{Theorem}[section]

\newtheorem{prop}[thm]{Proposition}

\newcommand{\term}[1]{\textbf{#1}}

\newcommand{\figref}[1]{Fig.~\ref{#1}}
\newcommand{\eqnref}[1]{Eq.~(\ref{#1})}
\newcommand{\secref}[1]{Sec.~\ref{#1}}

\newcommand{\algref}[1]{Alg.~\ref{#1}}

\newcommand{\Lone}{L_{1}}

\newcommand{\LOneNorm}[1]{\left|\left| #1 \right|\right|_1}

\newcommand{\set}[1]{\left\{#1\right\}}

\newcommand{\size}[1]{\left| #1 \right|}

\newcommand{\allverts}{V}

\newcommand{\alledges}{E}

\newcommand{\vertexdata}[1]{D_{#1}}
\newcommand{\vertexdatain}[1]{\edgedata{*}{#1}}
\newcommand{\vertexdataout}[1]{\edgedata{#1}{*}}
\newcommand{\edgedata}[2]{D_{#1 \rightarrow #2}}
\newcommand{\sdt}{\mathbf{T}}
\newcommand{\shareddata}[1]{\sdt\left[ #1 \right]}
\newcommand{\fold}{\text{Fold}}
\newcommand{\merge}{\text{Merge}}
\newcommand{\apply}{\text{Apply}}
\newcommand{\scope}[1]{\mathcal{S}_{#1}}

\SetKwInOut{Input}{Input}
\SetKwInOut{Output}{Output}
\SetKwFunction{Residual}{Residual}
\SetKwFor{ParForAll}{forall}{do in parallel}{}
\SetKwFunction{Reverse}{ReverseOrder}
\SetKwBlock{Sync}{Synchronized}{end}
\SetKwFunction{Map}{Map}
\SetKwFunction{Reduce}{Reduce}
\SetKwFunction{Push}{Push}
\SetKwFunction{InitializeQueue}{InitializeQueue}
\SetKwFunction{TopResid}{TopResid}
\SetKwFunction{Top}{Top}
\SetKwFunction{Pop}{Pop}
\SetKwFunction{Enqueue}{Enqueue}
\SetKwFunction{Dequeue}{Dequeue}
\SetKwFunction{UpdatePriority}{UpdatePriority}
\SetKwFunction{SetPriority}{SetPriority}
\SetKwFor{Lock}{lock}{}{release}
\SetKwFunction{Collect}{Collect}
\SetKwFunction{Initialize}{Initialize}
\SetKwFunction{TokenRing}{TokenRing}
\SetKwFunction{Promote}{Promote}
\SetKwFunction{Value}{Value}

\title{GraphLab: A New Framework For Parallel Machine Learning}

\author{
  {\bf Yucheng Low} \\
  Carnegie Mellon University \\
  ylow@cs.cmu.edu \\
  \And
  {\bf Joseph Gonzalez } \\
  Carnegie Mellon University \\
  jegonzal@cs.cmu.edu \\ 
  \And
  {\bf Aapo Kyrola } \\
  Carnegie Mellon University \\
  akyrola@cs.cmu.edu \\ 
  \AND
  {\bf Danny Bickson} \\
  Carnegie Mellon University \\
  bickson@cs.cmu.edu \\ 
  \And
  {\bf Carlos Guestrin} \\
  Carnegie Mellon University \\
  guestrin@cs.cmu.edu \\
  \And
  {\bf Joseph M. Hellerstein} \\
  UC Berkeley \\
  hellerstein@cs.berkeley.edu \\ 
}

\begin{document}

\maketitle

\begin{abstract}
  Designing and implementing \emph{efficient}, \emph{provably correct}
  parallel machine learning (ML) algorithms is challenging.  Existing
  high-level parallel abstractions like MapReduce are insufficiently
  expressive while low-level tools like MPI and Pthreads leave ML
  experts repeatedly solving the same design challenges.  By targeting
  common patterns in ML, we developed GraphLab, which improves upon
  abstractions like MapReduce by compactly expressing asynchronous
  iterative algorithms with sparse computational dependencies while
  ensuring data consistency and achieving a high degree of parallel
  performance.  We demonstrate the expressiveness of the GraphLab
  framework by designing and implementing parallel versions of belief
  propagation, Gibbs sampling, Co-EM, Lasso and Compressed Sensing.
  We show that using GraphLab we can achieve excellent parallel
  performance on large scale real-world problems.
\end{abstract}



\tightsection{Introduction}

Exponential gains in hardware technology have enabled sophisticated
machine learning (ML) techniques to be applied to increasingly
challenging real-world problems.  However, recent developments in
computer architecture have shifted the focus away from frequency
scaling and towards parallel scaling, threatening the future of
sequential ML algorithms.  In order to benefit from future trends in
processor technology and to be able to apply rich structured models to
rapidly scaling real-world problems, the ML community must directly
confront the challenges of parallelism.
 

However, designing and implementing \emph{efficient} and
\emph{provably correct} parallel algorithms is extremely
challenging. While low level abstractions like MPI and Pthreads
provide powerful, expressive primitives, they force the user to
address hardware issues and the challenges of parallel data
representation.  Consequently, many ML experts have turned to
high-level abstractions, which dramatically simplify the design and
implementation of a \emph{restricted} class of parallel algorithms.
For example, the MapReduce abstraction \citep{dean04} has been
successfully applied to a broad range of ML applications
\citep{cheng06, wolfe08, panda09, ye09}.

However, by restricting our focus to ML algorithms that are naturally
expressed in MapReduce, we are often forced to make overly simplifying
assumptions. Alternatively, by coercing efficient sequential ML
algorithms to satisfy the restrictions imposed by MapReduce, we often
produce inefficient parallel algorithms that require many processors
to be competitive with comparable sequential methods.

In this paper we propose GraphLab, a new parallel framework for ML
which exploits the \emph{sparse structure} and common
\emph{computational patterns} of ML algorithms.  GraphLab
enables ML experts to easily design and implement efficient scalable
parallel algorithms by composing problem specific computation,
data-dependencies, and scheduling.  We provide an efficient
\emph{shared-memory} implementation\footnote{The C++ reference
  implementation of the GraphLab is available at
  \url{http://select.cs.cmu.edu/code}.  } of GraphLab and use it to
build parallel versions of four popular ML algorithms.  We focus on
the shared-memory multiprocessor setting because it is both ubiquitous
and has few effective high-level abstractions.  We evaluate the
algorithms on a $16$-processor system and demonstrate state-of-the-art
performance. \emph{Our main contributions include}: \vspace{-0.2cm}
\begin{itemize}
  \lessdenselist
\item A graph-based data model which simultaneously represents data
  and computational dependencies.
\item A set of concurrent access models which provide a range of
  sequential-consistency guarantees.
\item A sophisticated modular scheduling mechanism.
\item An aggregation framework to manage global state.
\item GraphLab implementations and experimental evaluations of
  parameter learning and inference in graphical models, Gibbs
  sampling, CoEM, Lasso and compressed sensing on real-world problems.
\end{itemize}

\tightsection{Existing Frameworks}
\label{sec:otherframeworks}

There are several existing frameworks for designing and implementing
parallel ML algorithms.  Because GraphLab generalizes these ideas and
addresses several of their critical limitations we briefly review these
frameworks.




\tightsubsection{ Map-Reduce Abstraction }

A program implemented in the MapReduce framework consists of a \Map
operation and a \Reduce operation. The \Map operation is a function
which is applied independently and in parallel to each datum (e.g.,
webpage) in a large data set (e.g., computing the word-count).  The
\Reduce operation is an aggregation function which combines the \Map
outputs (e.g., computing the total word count).  MapReduce performs
optimally only when the algorithm is \emph{embarrassingly parallel}
and can be decomposed into a large number of independent computations.
The MapReduce framework expresses the class of ML algorithms which fit
the Statistical-Query model \citep{cheng06} as well as problems where
feature extraction dominates the run-time.

The MapReduce abstraction fails when there are \emph{computational
  dependencies} in the data.  For example, MapReduce can be used to
extract features from a massive collection of images but cannot
represent computation that depends on small overlapping subsets of
images.  This critical limitation makes it difficult to represent
algorithms that operate on structured models.  As a consequence, when
confronted with large scale problems, we often abandon rich structured
models in favor of overly simplistic methods that are amenable to the
MapReduce abstraction.


Many ML algorithms \emph{iteratively} transform parameters during both
learning and inference.  For example, algorithms like Belief
Propagation (BP), EM, gradient descent, and even Gibbs sampling,
iteratively refine a set of parameters until some termination
condition is achieved. While the MapReduce abstraction can be invoked
iteratively, it does not provide a mechanism to directly encode
iterative computation. As a consequence, it is not possible to
express sophisticated scheduling, automatically assess termination, or
even leverage basic data persistence.

The popular implementations of the MapReduce abstraction are targeted
at large data-center applications and therefore optimized to address
node-failure and disk-centric parallelism.  The overhead associated
with the fault-tolerant, disk-centric approach is unnecessarily costly
when applied to the typical cluster and multi-core settings
encountered in ML research.  Nonetheless, MapReduce is used in small
clusters and even multi-core settings \citep{cheng06}.  The GraphLab
implementation\footnote{The GraphLab abstraction is intended for both
  the multicore and cluster settings and a distributed, fault-tolerant
  implementation is ongoing research.} described in this paper does
not address fault-tolerance or parallel disk access and instead
assumes that processors do not fail and all data is stored in
shared-memory.  As a consequence, GraphLab does not incur the
unnecessary disk overhead associated with MapReduce in the multi-core
setting.

\tightsubsection{DAG Abstraction} In the DAG abstraction, parallel
computation is represented as a directed acyclic graph with data
flowing along edges between vertices. Vertices correspond to functions
which receive information on inbound edges and output results to
outbound edges.  Implementations of this abstraction include Dryad
\citep{dryad} and Pig Latin \citep{piglatin}.

While the DAG abstraction permits rich computational dependencies it
does not naturally express iterative algorithms since the structure of
the dataflow graph depends on the number of iterations (which must
therefore be known prior to running the program).  The DAG abstraction
also cannot express dynamically prioritized computation.

\tightsubsection{Systolic Abstraction}

The Systolic abstraction \citep{systolicvlsi} (and the closely related
Dataflow abstraction) extends the DAG framework to the iterative
setting. Just as in the DAG Abstraction, the Systolic abstraction
forces the computation to be decomposed into small atomic components
with limited communication between the components.  The Systolic
abstraction uses a directed graph $G=(\allverts,\alledges)$ which is
not necessarily acyclic) where each vertex represents a processor, and
each edge represents a communication link. In a single iteration, each
processor reads all incoming messages from the in-edges, performs some
computation, and writes messages to the out-edges. A barrier
synchronization is performed between each iteration, ensuring all
processors compute and communicate in lockstep.

%

While the Systolic framework can express iterative computation, it is
unable to express the wide variety of update schedules used in ML
algorithms. For example, while gradient descent may be run within the
Systolic abstraction using a \term{Jacobi schedule} it is not possible to
implement coordinate descent which requires the more sequential
\term{Gauss-Seidel schedule}.  The Systolic abstraction also cannot express
the dynamic and specialized structured schedules which were shown by
\cite{aistats, gonzalez09} to dramatically improve the performance of
algorithms like BP.

\tightsection{The GraphLab Abstraction}

By targeting common patterns in ML, like sparse data dependencies and
asynchronous iterative computation, GraphLab achieves a balance
between low-level and high-level abstractions. Unlike many low-level
abstractions (e.g., MPI, PThreads), GraphLab insulates users from the
complexities of synchronization, data races and deadlocks by providing
a high level data representation through the \term{data graph} and
automatically maintained data-consistency guarantees through
configurable \term{consistency models}.  Unlike many high-level
abstractions (i.e., MapReduce), GraphLab can express complex
computational dependencies using the \term{data graph} and provides
sophisticated \term{scheduling primitives} which can express iterative
parallel algorithms with dynamic scheduling.

To aid in the presentation of the GraphLab framework we use Loopy
Belief Propagation (BP) \citep{pearl88} on pairwise Markov Random
Fields (MRF) as a running example.  A pairwise MRF is an undirected
graph over random variables where edges represent interactions between
variables. Loopy BP is an approximate inference algorithm which
estimates the marginal distributions by iteratively recomputing
parameters (messages) associated with each edge until some convergence
condition is achieved.  






%
%

\tightsubsection{Data Model}

The GraphLab data model consists of two parts: a directed \term{data
  graph} and a \term{shared data table}. The data graph $G=(V,E)$
encodes both the problem specific \emph{sparse computational
  structure} and directly modifiable program state.  The user can
associate arbitrary blocks of data (or parameters) with each vertex
and directed edge in $G$.  We denote the data associated with vertex
$v$ by $\vertexdata{v}$, and the data associated with edge $(u
\rightarrow v)$ by $\edgedata{u}{v}$. In addition, we use
$(u\rightarrow *)$ to represent the set of all outbound edges from $u$
and $(*\rightarrow v)$ for inbound edges at $v$.  To support globally
shared state, GraphLab provides a \term{shared data table} (SDT) which
is an associative map, $\shareddata{\text{Key}} \rightarrow
\text{Value}$, between keys and arbitrary blocks of data.




In the Loopy BP, the data graph is the pairwise MRF, with the vertex
data $\vertexdata{v}$ to storing the node potentials and the directed
edge data $\edgedata{u}{v}$ storing the BP messages.  If the MRF is
sparse then the data graph is also sparse and GraphLab will achieve a
high degree of parallelism.  The SDT can be used to store shared
hyper-parameters and the global convergence progress.


\tightsubsection{User Defined Computation}

Computation in GraphLab can be performed either through an
\term{update function} which defines the local computation, or through
the \term{sync mechanism} which defines global aggregation. The Update
Function is analogous to the \Map in MapReduce, but unlike in
MapReduce, update unctions are permitted to access and modify
\emph{overlapping} contexts in the graph. The sync mechanism is
analogous to the \Reduce operation, but unlike in MapReduce, the sync
mechanism runs concurrently with the update functions.



\subsubsection{Update Functions}

A GraphLab \term{update function} is a stateless user-defined function
which operates on the data associated with small neighborhoods in the
graph and represents the core element of computation.  For every
vertex $v$, we define $\scope{v}$ as the neighborhood of $v$ which
consists of $v$, its adjacent edges (both inbound and outbound) and
its neighboring vertices as shown in \figref{fig:scope}.  We define
$\vertexdata{\scope{v}}$ as the data corresponding to the neighborhood
$\scope{v}$. In addition to $\vertexdata{\scope{v}}$, update functions
also have \emph{read-only} access, to the shared data table $\sdt$.
We define the application of the update function $f$ to the vertex $v$
as the state mutating computation:
\[
\vertexdata{\scope{v}} \leftarrow
f(\vertexdata{\scope{v}}, \sdt).
\]
We refer to the neighborhood $\scope{v}$ as the \term{scope} of $v$
because $\scope{v}$ defines the extent of the graph that can be
accessed by $f$ when applied to $v$.  For notational simplicity, we
denote $f(\vertexdata{\scope{v}}, \sdt)$ as $f(v)$. A GraphLab program
may consist of multiple update functions and it is up to the
scheduling model (see \secref{sec:scheduling}) to determine which
update functions are applied to which vertices and in which parallel
order.

\subsubsection{Sync Mechanism}
\label{sec:sync}
\begin{algorithm}[t]
  \footnotesize
  \label{alg:sync}
  \caption{Sync Algorithm on $k$}
  \dontprintsemicolon
  $t \leftarrow r_{k}^{(0)}$  \;
  \ForEach{$v \in \allverts$ } {
    $t \leftarrow \fold_{k}(\vertexdata{v}, t)$ \;
  }
  $\shareddata{k} \leftarrow \apply_{k}(t) $   \;
\end{algorithm}

The \term{sync mechanism} aggregates data across all vertices in the
graph in a manner analogous to the Fold and Reduce operations in
functional programming. The result of the sync operation is associated
with a particular entry in the Shared Data Table (SDT). The user
provides a key $k$, a \term{fold function} (\eqnref{eqn:sync}), an
\term{apply function} (\eqnref{eqn:apply}) as well as an initial value
$r_{k}^{(0)}$ to the SDT and an optional \term{merge function} used
to construct parallel tree reductions.
\begin{eqnarray}
  r_k^{(i+1)} & \leftarrow &
  \fold_{k}\left(\vertexdata{v}, r_k^{(i)} \right)
  \label{eqn:sync} \\
  r_k^{l} & \leftarrow &
  \merge_{k}\left(r_k^{i}, r_k^{j} \right)
  \label{eqn:reduce} \\
  \shareddata{k} & \leftarrow & \apply_{k}(r_{k}^{(\size{\allverts})}) 
  \label{eqn:apply}
\end{eqnarray}
When the sync mechanism is invoked, the algorithm in \algref{alg:sync}
uses the $\fold_{k}$ function to \emph{sequentially} aggregate data
across all vertices.  The $\fold_{k}$ function obeys the same
consistency rules (described in \secref{sec:consistency}) as update
functions and is therefore able to \emph{modify} the vertex data.  If
the $\merge_{k}$ function is provided a \emph{parallel} tree reduction
is used to combine the results of multiple parallel folds. The
$\apply_{k}$ then finalizes the resulting value (e.g., rescaling)
before it is written back to the SDT with key $k$.

The sync mechanism can be set to run periodically in the background
while the GraphLab engine is actively applying update functions or on
demand triggered by update functions or user code.  If the sync
mechanism is executed in the background, the resulting aggregated
value may not be globally consistent.  Nonetheless, many ML
applications are robust to approximate global statistics.


In the context of the Loopy BP example, the update function is the BP
message update in which each vertex recomputes its outbound messages
by integrating the inbound messages.  The sync mechanism is used to
monitor the global convergence criterion (for instance, average change
or residual in the beliefs).  The $\fold_{k}$ function accumulates the
residual at the vertex, and the $\apply_{k}$ function divides the
final answer by the number of vertices.  To monitor progress, we let
GraphLab run the sync mechanism as a periodic background process.



\begin{figure}[t]
  \begin{center}
    \subfigure[Scope]{
      \label{fig:scope}
      \includegraphics[width=.45\textwidth]{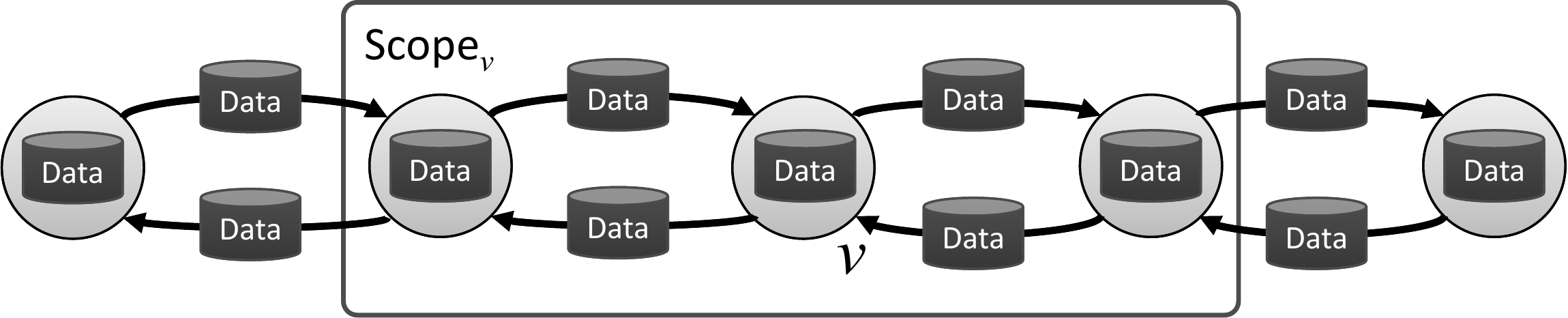}
    }
    \subfigure[Consistency Models]{
      \label{fig:locking}
      \includegraphics[width=.45\textwidth]{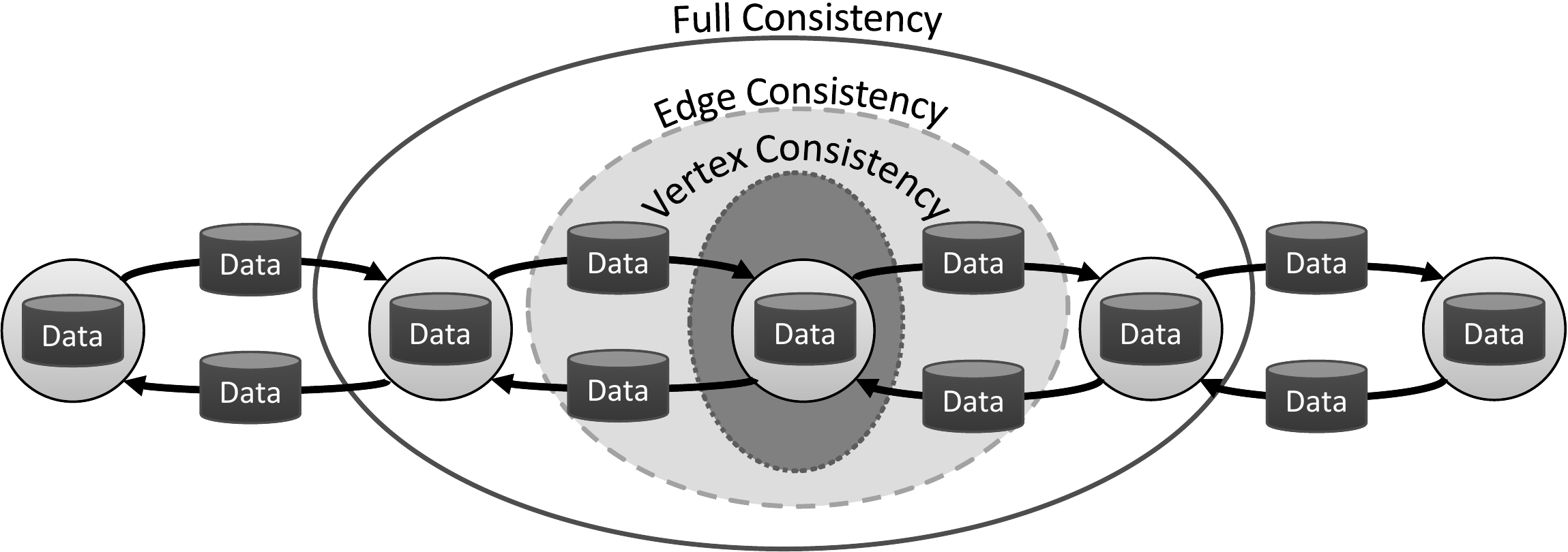}
    }
    \caption{ \footnotesize \textbf{(a)} The scope, $\scope{v}$, of
      vertex $v$ consists of all the data at the vertex $v$, its
      inbound and outbound edges, and its neighboring vertices.  The
      update function $f$ when applied to the vertex $v$ can read and
      modify any data within $\scope{v}$. \textbf{(b)}. We illustrate
      the 3 data consistency models by drawing their exclusion sets as
      a ring where no two update functions may be executed
      simultaneously if their exclusions sets (rings) overlap.}
  \end{center}
  \vspace{-0.5cm}
\end{figure}


\tightsubsection{Data Consistency}
\label{sec:consistency}

Since scopes may overlap, the simultaneous execution of two update
functions can lead to race-conditions resulting in data inconsistency
and even corruption.  For example, two function applications to
neighboring vertices could simultaneously try to modify data on a
shared edge resulting in a corrupted value.  Alternatively, a function
trying to normalize the parameters on a set of edges may compute the
sum only to find that the edge values have changed.

GraphLab provides a choice of three data consistency models which
enable the user to balance performance and data consistency.  The
choice of data consistency model determines the extent to which
overlapping scopes can be executed simultaneously. We illustrate each
of these models in \figref{fig:locking} by drawing their corresponding
\term{exclusion sets}.  GraphLab guarantees that update functions
never simultaneously share overlapping exclusion sets. Therefore
larger exclusion sets lead to reduced parallelism by delaying the
execution of update functions on nearby vertices.




The \term{full consistency} model ensures that during the execution of
$f(v)$ no other function will read or modify data within
$\scope{v}$. Therefore, parallel execution may only occur on vertices
that do not share a common neighbor. The slightly weaker \term{edge
  consistency} model ensures that during the execution of $f(v)$ no
other function will read or modify any of the data on $v$ or any of 
the edges adjacent to $v$.  Under the edge consistency model, parallel 
execution may only occur on non-adjacent vertices.  Finally, the weakest
\term{vertex consistency} model only ensures that during the execution
of $f(v)$ no other function will be applied to $v$. The vertex
consistency model is therefore prone to race conditions and should
only be used when reads and writes to adjacent data can be done 
safely (In particular repeated reads may return different results). However, by
permitting update functions to be applied simultaneously to
neighboring vertices, the vertex consistency model permits maximum
parallelism.



Choosing the right consistency model has direct implications to
program correctness.  One method to prove correctness of a parallel
algorithm is to show that it is equivalent to a correct sequential
algorithm.  To capture the relation between sequential and parallel
execution of a program we introduce the concept of \term{sequential
  consistency}:
\begin{dfn}[Sequential Consistency]
  A GraphLab program is \term{sequentially consistent} if for every
  parallel execution, there exists a sequential execution of update
  functions that produces an equivalent result.
\end{dfn}
The sequential consistency property is typically a sufficient
condition to extend algorithmic correctness from the sequential
setting to the parallel setting. In particular, if the algorithm is
correct under \emph{any} sequential execution of update functions, then the
parallel algorithm is also correct if sequential consistency is
satisfied.
\begin{prop}
  GraphLab guarantees sequential consistency under the following three
  conditions:
   \begin{enumerate}
     \vspace{-10pt}
     \lessdenselist
   \item The \term{full consistency} model is used
   \item The \term{edge consistency} model is used and update
     functions do not modify data in adjacent vertices.
   \item The \term{vertex consistency} model is used and update
     functions only access local vertex data.
   \end{enumerate}
\end{prop}

In the Loopy BP example the update function only needs to read and
modify data on the adjacent edges. Therefore the edge consistency
model ensures sequential consistency.
%
%

\tightsubsection{Scheduling}
\label{sec:scheduling}
The GraphLab \term{update schedule} describes the order in which
update functions are applied to vertices and is represented by a
parallel data-structure called the \term{scheduler}.  The
\term{scheduler} abstractly represents a dynamic list of \term{tasks}
(vertex-function pairs) which are to be executed by the GraphLab
\term{engine}.

Because constructing a scheduler requires reasoning about the
complexities of parallel algorithm design, the GraphLab framework
provides a collection of base schedules.  To represent Jacobi style
algorithms (e.g., gradient descent) GraphLab provides a
\term{synchronous scheduler} which ensures that all vertices are
updated simultaneously.  To represent Gauss-Seidel style algorithms
(e.g., Gibbs sampling, coordinate descent), GraphLab provides a
\term{round-robin scheduler} which updates all vertices
\emph{sequentially} using the most recently available data.

Many ML algorithms (e.g., Lasso, CoEM, Residual BP) require more
control over the tasks that are created and the order in which they
are executed. Therefore, GraphLab provides a collection of \term{task
  schedulers} which permit update functions to add and reorder tasks.
GraphLab provides two classes of task schedulers.  The \term{FIFO}
schedulers only permit task creation but do not permit task
reordering.  The \term{prioritized} schedules permit task reordering
at the cost of increased overhead.  For both types of task scheduler
GraphLab also provide relaxed versions which increase performance at
the expense of reduced control:
\begin{center}
\footnotesize
\begin{tabular}{|l|l|l|}
\hline
& \textbf{Strict Order} & \textbf{Relaxed Order}    \\
\hline \hline
\textbf{FIFO}        & Single Queue    & Multi Queue / Partitioned \\
\textbf{Prioritized} & Priority Queue  & Approx. Priority Queue \\
\hline
\end{tabular}
\end{center}
In addition GraphLab provides the \term{splash scheduler} based on the
loopy BP schedule proposed by \cite{aistats} which executes tasks
along spanning trees.

In the Loopy BP example, different choices of scheduling leads to
different BP algorithms. Using the Synchronous scheduler corresponds
to the classical implementation of BP and using priority scheduler
corresponds to Residual BP \citep{elidan06}.

\subsubsection{Set Scheduler}
\label{sec:setscheduler}

Because scheduling is important to parallel algorithm design, GraphLab
provides a scheduler construction framework called the \term{set
  scheduler} which enables users to \emph{safely} and \emph{easily}
compose custom update schedules.  To use the set scheduler the user
specifies a sequence of vertex set and update function pairs $((S_1,
f_1), (S_2, f_2) \cdots (S_k, f_k))$, where $S_i \subseteq V$ and
$f_i$ is an update function. This sequence implies the following
execution semantics:

\For{$i = 1\cdots k$} {
  Execute $f_i$ on all vertices in $S_i$ in parallel. \\
  Wait for all updates to complete 
}


The amount of parallelism depends on the size of each set; the
procedure is highly sequential if the set sizes are small.  Executing
the schedule in the manner described above can lead to the majority of
the processors waiting for a few processors to complete the current
set. However, by leveraging the causal data dependencies encoded in
the graph structure we are able to construct an \term{execution plan}
which identifies tasks in future sets that can be executed
\emph{early} while still producing an equivalent result.

\begin{figure}[t]
  \begin{center}
    \includegraphics[width=.48\textwidth]{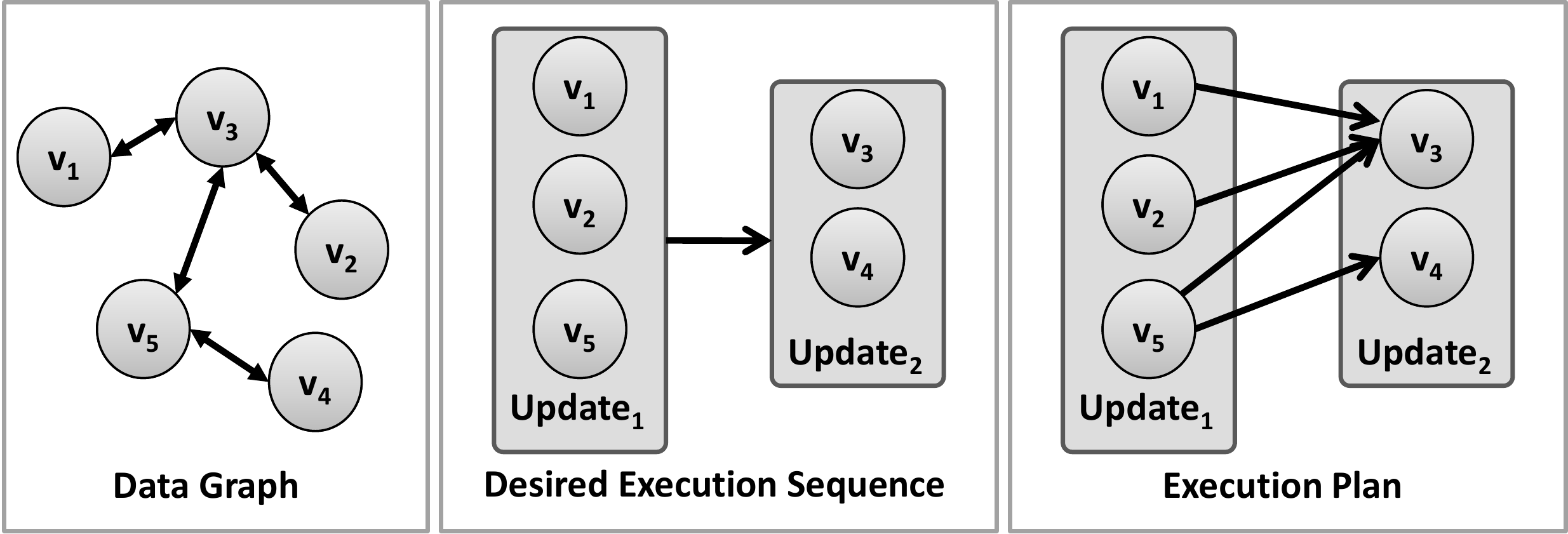}
    \caption{\footnotesize A simple example of the set scheduler
      planning process. Given the data graph, and a desired sequence
      of execution where $v_1, v_2$ and $v_5$ are first run in
      parallel, then followed by $v_3$ and $v_4$.  If the edge
      consistency model is used, we observe that the execution of
      $v_3$ depends on the state of $v_1, v_2$ and $v_5$, but the
      $v_4$ only depends on the state of $v_5$. The dependencies are
      encoded in the execution plan on the right. The resulting plan
      allows $v_4$ to be immediately executed after $v_5$ without
      waiting for $v_1$ and $v_2$.}
    \label{fig:setschedule}
  \end{center}
\end{figure}

The set scheduler compiles an execution plan by rewriting the
execution sequence as a Directed Acyclic Graph (DAG), where each
vertex in the DAG represents an update task in the execution sequence
and edges represent execution dependencies.  \figref{fig:setschedule}
provides an example of this process.  The DAG imposes a partial
ordering over tasks which can be compiled into a parallel execution
schedule using the greedy algorithm described by
\cite{Graham66}. 



\begin{figure}[t]
  \begin{center}
    \includegraphics[width=.45\textwidth]{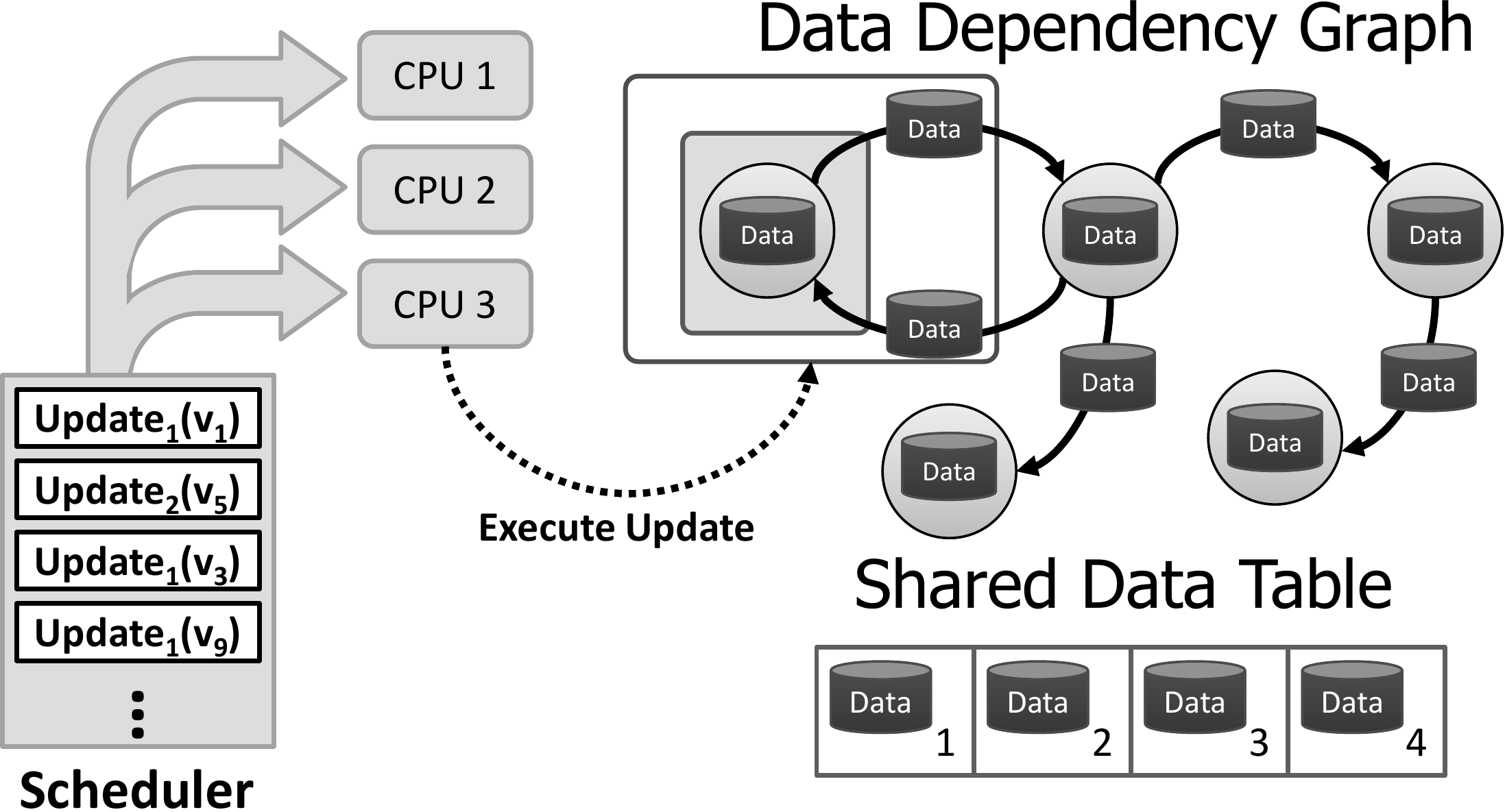}
    \caption{ \footnotesize A summary of the GraphLab framework. The
      user provides a graph representing the computational data
      dependencies, as well as a SDT containing read only data. The
      user also picks a scheduling method or defines a custom
      schedule, which provides a stream of update tasks in the form of
      (vertex, function) pairs to the processors. }
    \label{fig:summary}
  \end{center}
  \vspace{-0.5cm}
\end{figure}

\tightsubsection{Termination Assessment}

Efficient parallel termination assessment can be challenging.  The
standard termination conditions used in many iterative ML algorithms
require reasoning about the global state.  The GraphLab framework
provides two methods for termination assessment. The first method
relies on the scheduler which signals termination when there are no
remaining tasks.  This method works for algorithms like Residual BP,
which use task schedulers and stop producing new tasks when they
converge.  The second termination method relies on user provided
termination functions which examine the SDT and signal when the
algorithm has converged.  Algorithms, like parameter learning, which
rely on global statistics use this method.


\tightsubsection{Summary and Implementation}
A GraphLab program is composed of the following parts:
\begin{enumerate}
  \vspace{-8pt}
  \lessdenselist
\item A \term{data graph} which represents the data and computational
  dependencies.
\item \term{Update functions} which describe local computation 
\item A \term{Sync mechanism} for aggregating global state.
\item A data \term{consistency model} (i.e., \emph{Fully Consistent},
  \emph{Edge Consistent} or \emph{Vertex Consistent}), which
  determines the extent to which computation can overlap.
\item \term{Scheduling primitives} which express the order of
  computation and may depend dynamically on the data.
\end{enumerate}

We implemented an optimized version of the GraphLab framework in C++
using PThreads.  The resulting GraphLab API is available under the
LGPL license at {\small \url{http://select.cs.cmu.edu/code}}.  The
data consistency models were implemented using race-free and
deadlock-free ordered locking protocols. To attain maximum performance
we addressed issues related to parallel memory allocation, concurrent
random number generation, and cache efficiency.  Since mutex
collisions can be costly, lock-free data structures and atomic
operations were used whenever possible.  To achieve the same level of
performance for parallel learning system, the ML community would have
to repeatedly overcome many of the same \emph{time consuming} systems
challenges needed to build GraphLab.

The GraphLab API has the opportunity to be an interface between the ML
and systems communities.  Parallel ML algorithms built around the
GraphLab API automatically benefit from developments in parallel data
structures.  As new locking protocols and parallel scheduling
primitives are incorporated into the GraphLab API, they become
immediately available to the ML community.  Systems experts can more
easily port ML algorithms to new parallel hardware by porting the
GraphLab API.


%
%
%

\tightsection{Case Studies}

To demonstrate the expressiveness of the GraphLab abstraction and
illustrate the parallel performance gains it provides, we implement
four popular ML algorithms and evaluate these algorithms on large
real-world problems using a 16-core computer with 4 AMD Opteron 8384
processors and 64GB of RAM.

\tightsubsection{MRF Parameter Learning}
\label{sec:eyeball}

To demonstrate how the various components of the GraphLab framework
can be assembled to build a complete ML ``pipeline,'' we use GraphLab
to solve a novel three-dimensional retinal image denoising task.  In
this task we begin with raw three-dimensional laser density estimates,
then use GraphLab to generate composite statistics, learn parameters
for a large three-dimensional grid pairwise MRF, and then finally
compute expectations for each voxel using Loopy BP. Each of these
tasks requires both local iterative computation and global aggregation
as well as several different computation schedules.

We begin by using the GraphLab data-graph to build a large (256x64x64)
three dimensional MRF in which each vertex corresponds to a voxel in
the original image.  We connect neighboring voxels in the 6 axis
aligned directions.  We store the density observations and beliefs in
the vertex data and the BP messages in the directed edge data.  As
shared data we store three global edge parameters which determine the
smoothing (accomplished using a Laplace similarity potentials) in each
dimension.  Prior to learning the model parameters, we first use the
GraphLab sync mechanism to compute axis-aligned averages as a proxy
for ``ground-truth'' smoothed images along each dimension. We then
performed simultaneous learning and inference in GraphLab by using the
background sync mechanism (\algref{alg:bpsync}) to aggregate inferred
model statistics and apply a gradient descent procedure.  To the best
of our knowledge, this is the first time graphical model parameter
learning and BP inference have been done concurrently.


\begin{algorithm}[t]
  \footnotesize
  \caption{BP update function}
  \label{alg:bpfunction}
  \dontprintsemicolon
  \SetLine
  BPUpdate($\vertexdata{v}, \vertexdatain{v}, 
  \vertexdataout{v} \in \scope{v}$)
  \Begin {

    Compute the local belief $b(x_v)$ using 
    $\set{ \vertexdatain{v} \vertexdata{v} }$  \;

    \ForEach{$(v \rightarrow t) \in (v \rightarrow *)$} { 

      Update $m_{v \rightarrow t}(x_t)$ using $\set{ \vertexdatain{v},
        \vertexdata{v} }$ and $\lambda_{\text{axis}(v t)}$ from the
      SDT. \;
 
      residual $\leftarrow \LOneNorm{m_{v \rightarrow t}(x_t) - 
        m_{v \rightarrow t}^{\text{old}}(x_t) }$ \;
      
      \If{ residual $ > $ Termination Bound} {
        AddTask($t$, residual) \; 
      } 
    }
  }
\end{algorithm}

\begin{algorithm}[t]
  \footnotesize
  \caption{Parameter Learning Sync}
  \label{alg:bpsync}
  \dontprintsemicolon
  \SetLine
  Fold(acc, vertex)
  \Begin {
    Return acc + image statistics on vertex
  }
  Apply(acc) \Begin {
    Apply gradient step to $\lambda$ using acc 
    and return $\lambda$ \; 
  }
\end{algorithm}

\begin{figure*}[t]
  \begin{center}
    \subfigure[Speedup] {
      \label{fig:eyeball_speedup}
      \includegraphics[height=.16\textwidth]{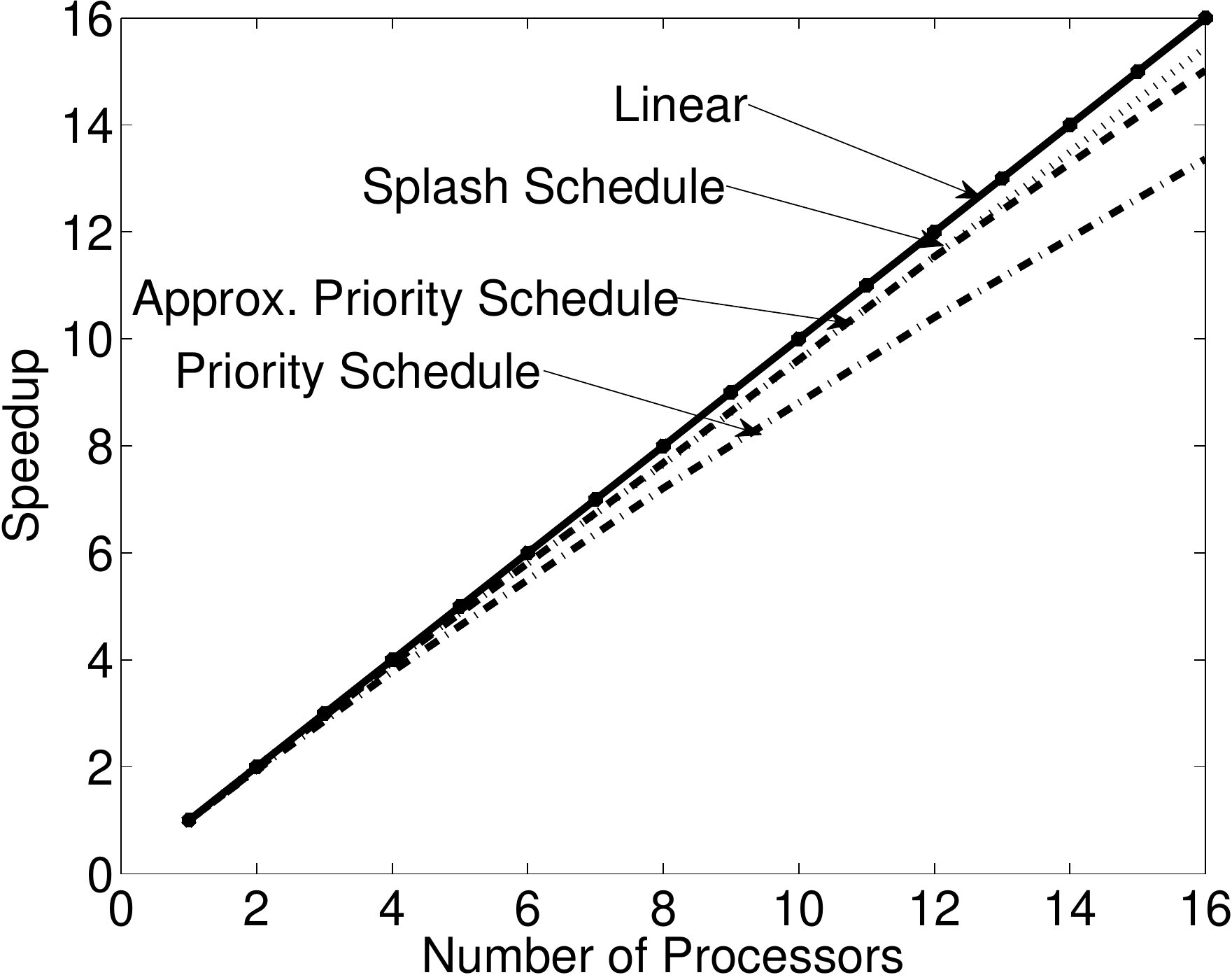}
    }
    \subfigure[Bkgnd Sync. Runtime] { 
      \label{fig:simul_lrn_inf_run}
      \includegraphics[height=.16\textwidth]{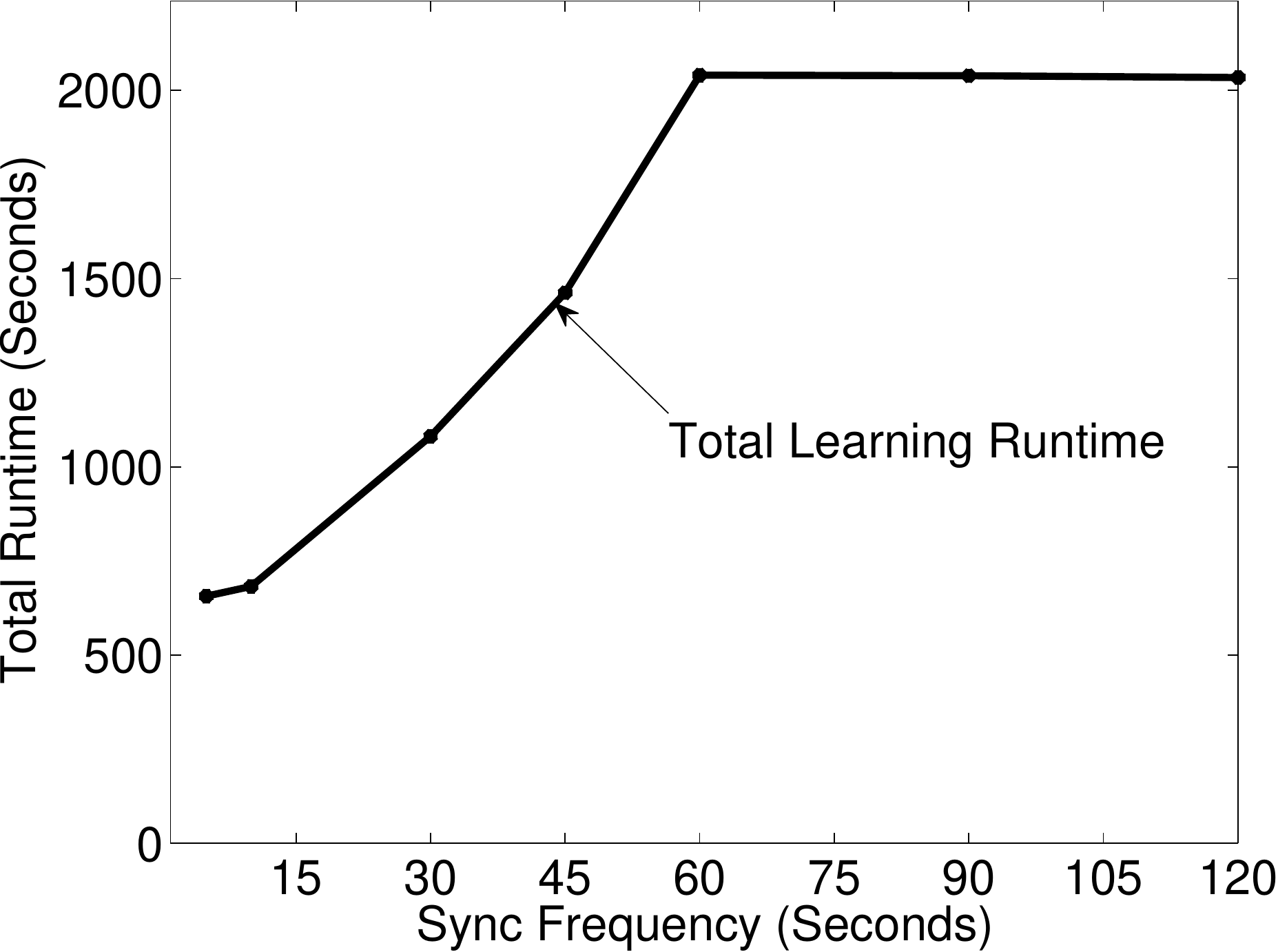}
    }
    \subfigure[Bkgnd Sync. Error] { 
      \label{fig:simul_lrn_inf_err}
      \includegraphics[height=.16\textwidth]{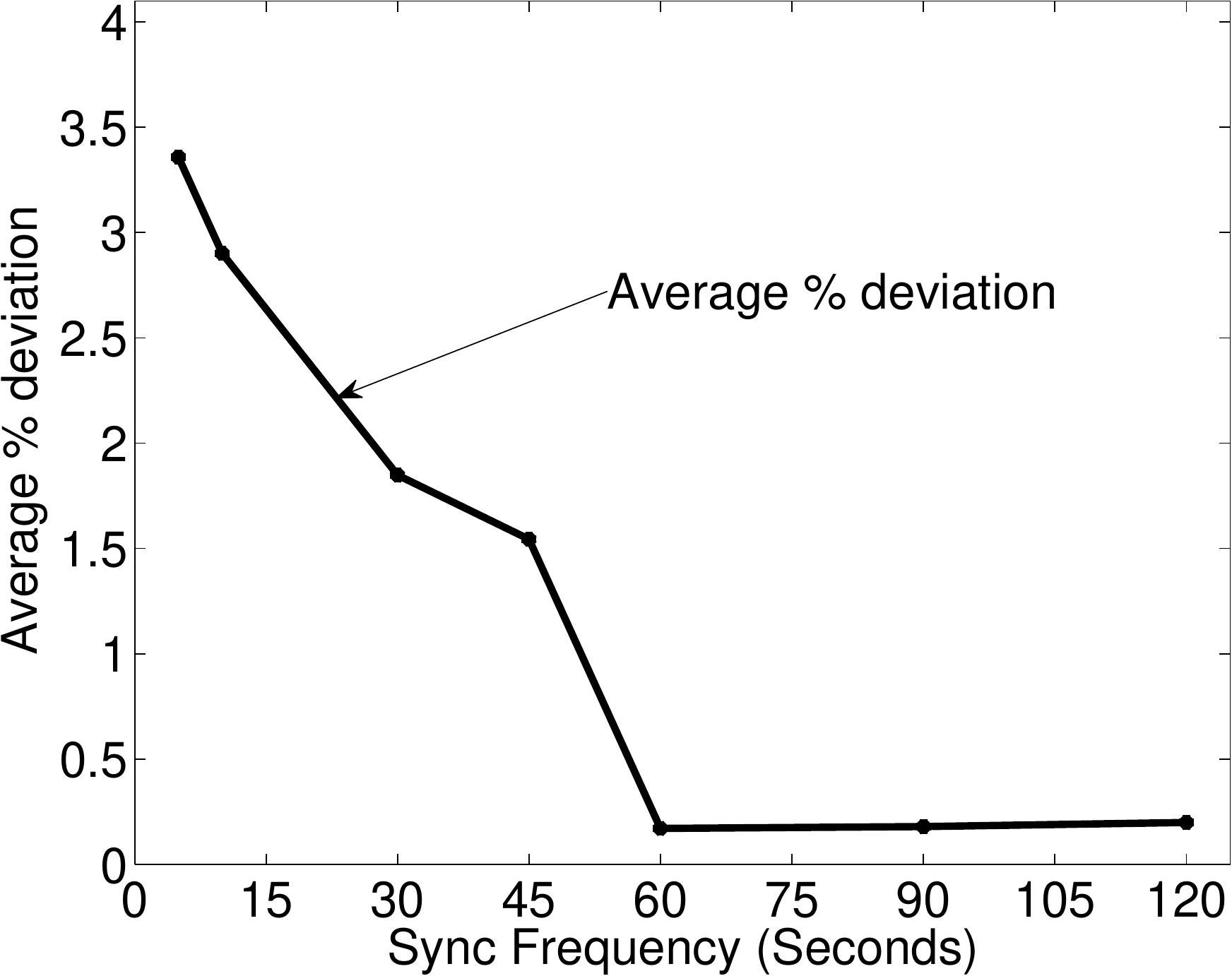}
    }
    \subfigure[Original] { 
      \label{fig:noisyeye}
      \includegraphics[height=.16\textwidth]{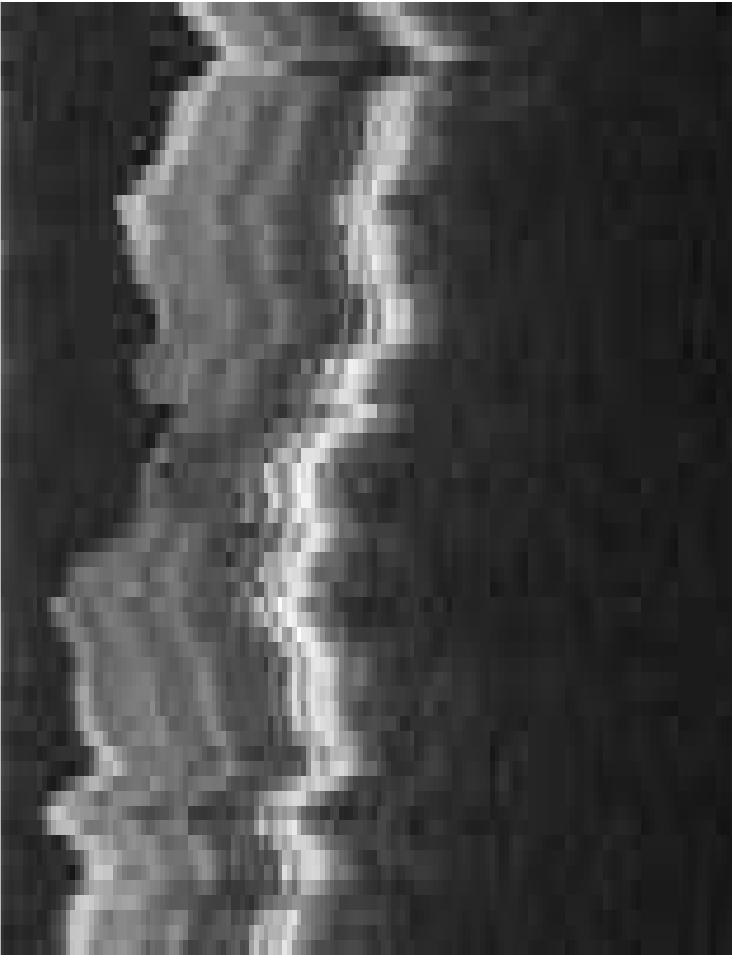}
    }
    \subfigure[Denoised] {
      \label{fig:cleaneye}
      \includegraphics[height=.16\textwidth]{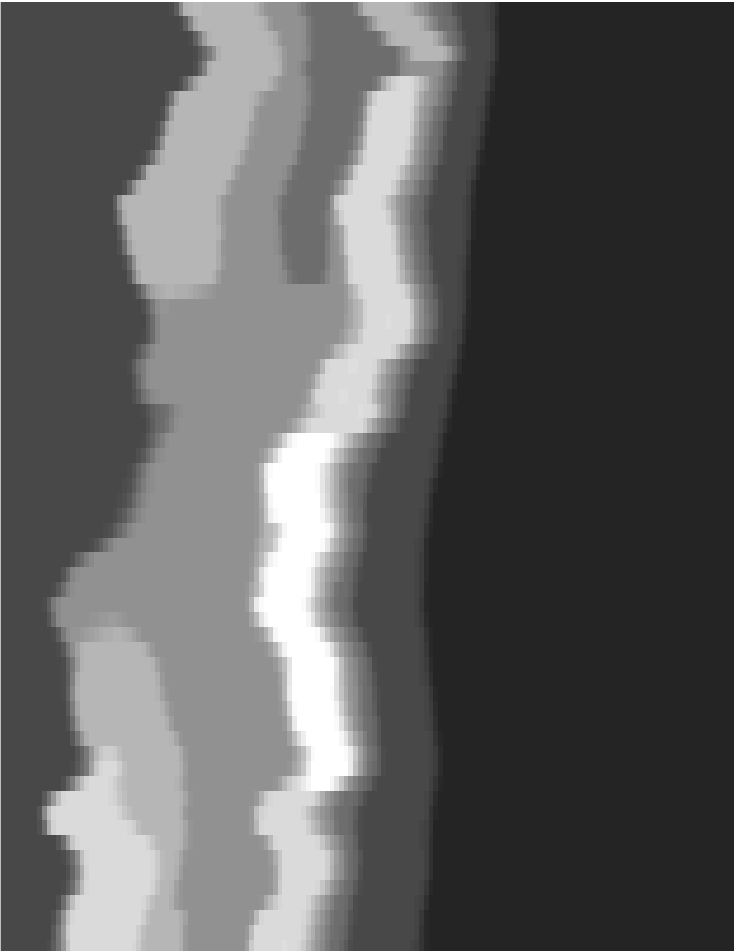}
    }
    \caption{\footnotesize \emph{Retinal Scan Denoising} \textbf{(a)}
      Speedup relative to the best single processor runtime of
      parameter learning using priority, approx priority, and Splash
      schedules.  \textbf{(b)} The total runtime in seconds of
      parameter learning and \textbf{(c)} the average percent
      deviation in learned parameters plotted against the time between
      gradient steps using the Splash schedule on 16 processors.
      \textbf{(d,e)} A slice of the original noisy image and the
      corresponding expected pixel values after parameter learning and
      denoising. }
  \end{center}

\end{figure*}

\paragraph{Results:} In \figref{fig:eyeball_speedup} we plot the
speedup of the parameter learning algorithm, executing inference and
learning sequentially.  We found that the Splash scheduler outperforms
other scheduling techniques enabling a factor 15 speedup on 16 cores.
We then evaluated simultaneous parameter learning and inference by
allowing the sync mechanism to run concurrently with inference
(\figref{fig:simul_lrn_inf_run} and \figref{fig:simul_lrn_inf_err}).
By running a background sync at the right frequency, we found that we
can further accelerate parameter learning while only marginally
affecting the learned parameters.  In \figref{fig:noisyeye} and
\figref{fig:cleaneye} we plot examples of noisy and denoised cross
sections respectively.

\tightsubsection{Gibbs Sampling}
\label{sec:gibbs}

The Gibbs sampling algorithm is inherently sequential and has
frustrated efforts to build asymptotically consistent parallel
samplers.  However, a standard result in parallel algorithms
\citep{bertsekas} is that for any fixed length Gauss-Seidel schedule
there exists an equivalent parallel execution which can be derived
from a coloring of the dependency graph.  We can extract this form of
parallelism using the GraphLab framework. We first use GraphLab to
construct a greedy graph coloring on the MRF and then to execute an
exact parallel Gibbs sampler.


We implement the standard greedy graph coloring algorithm in GraphLab
by writing an update function which examines the colors of the
neighboring vertices of $v$, and sets $v$ to the first unused
color. We use the edge consistency model with the parallel coloring
algorithm to ensure that the parallel execution retains the same
guarantees as the sequential version.  The parallel Gauss-Seidel
schedule is then built using the GraphLab set scheduler
(\secref{sec:setscheduler}) and the coloring of the MRF.  The
resulting schedule consists of a sequence of vertex sets $S_1$ to
$S_C$ such that $S_i$ contains all the vertices with color $i$.  
The \emph{vertex consistency} model is sufficient since the coloring
ensures full sequential consistency.

To evaluate the GraphLab parallel Gibbs sampler we consider the
challenging task of marginal estimation on a factor graph representing
a protein-protein interaction network obtained from \cite{elidan06} by
generating $10,000$ samples. The resulting MRF has roughly $100K$
edges and $14K$ vertices. As a baseline for comparison we also ran a
GraphLab version of the highly optimized Splash Loopy BP
\citep{gonzalez09} algorithm.

\begin{figure*}
   \begin{center}
     \subfigure[Gibbs Speedup] {
       \label{fig:gibbsspeedup}
       \includegraphics[width=.185\textwidth]{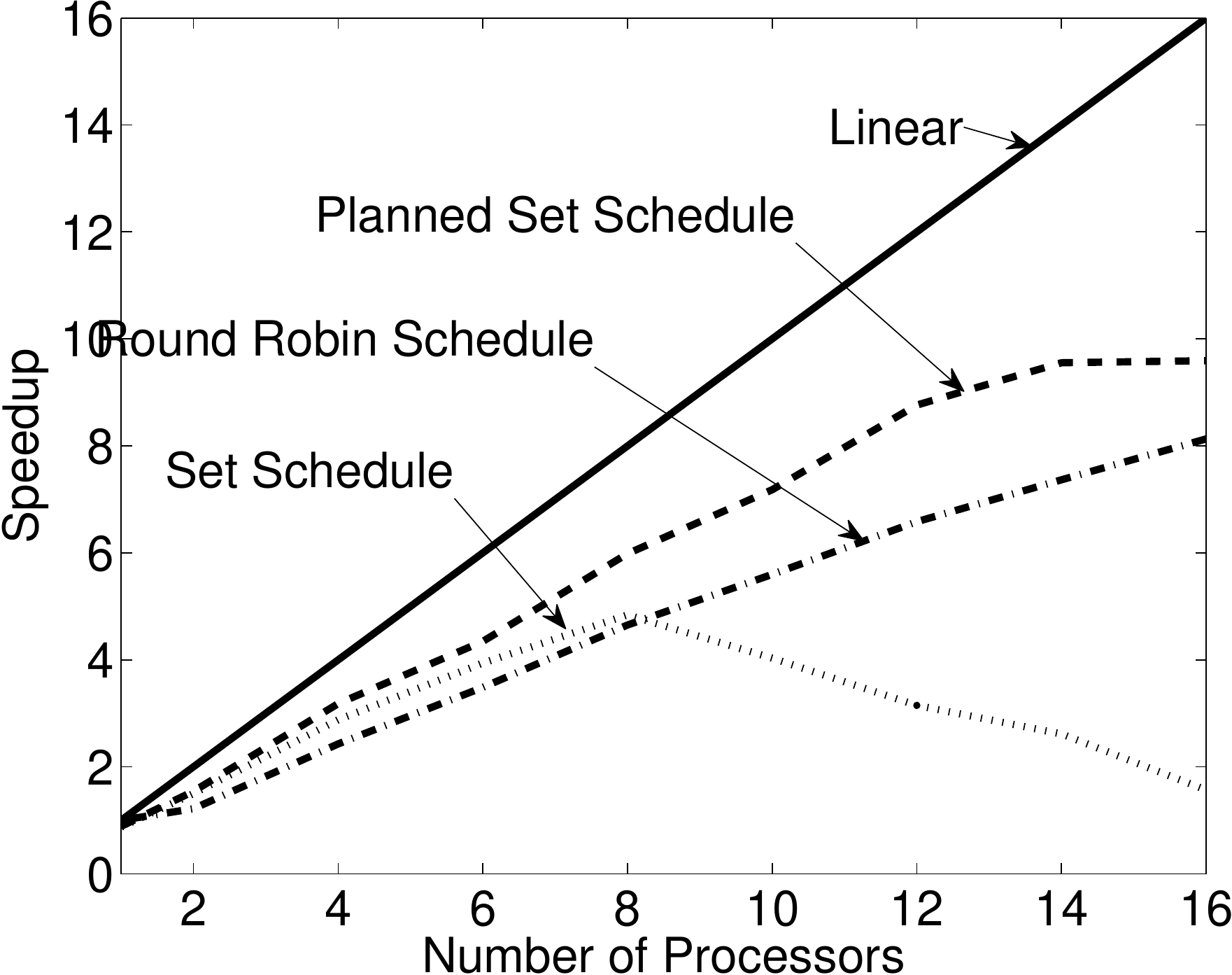}
     }
     \subfigure[Gibbs Color] { 
       \label{fig:gibbscolor}
       \includegraphics[width=.19\textwidth]{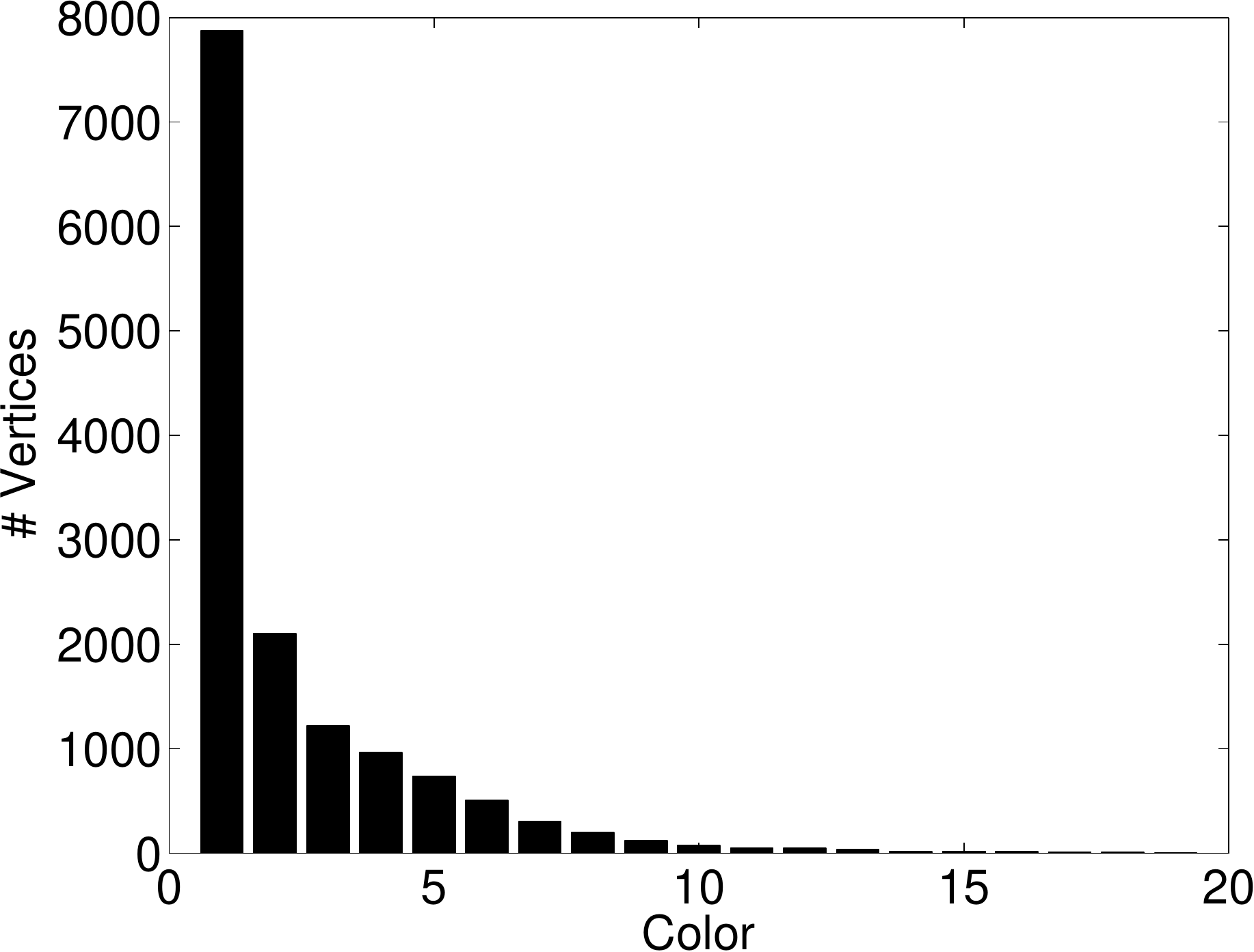}
     }
     \subfigure[Gibbs Eff.] { 
       \label{fig:gibbseff}
       \includegraphics[width=.185\textwidth]{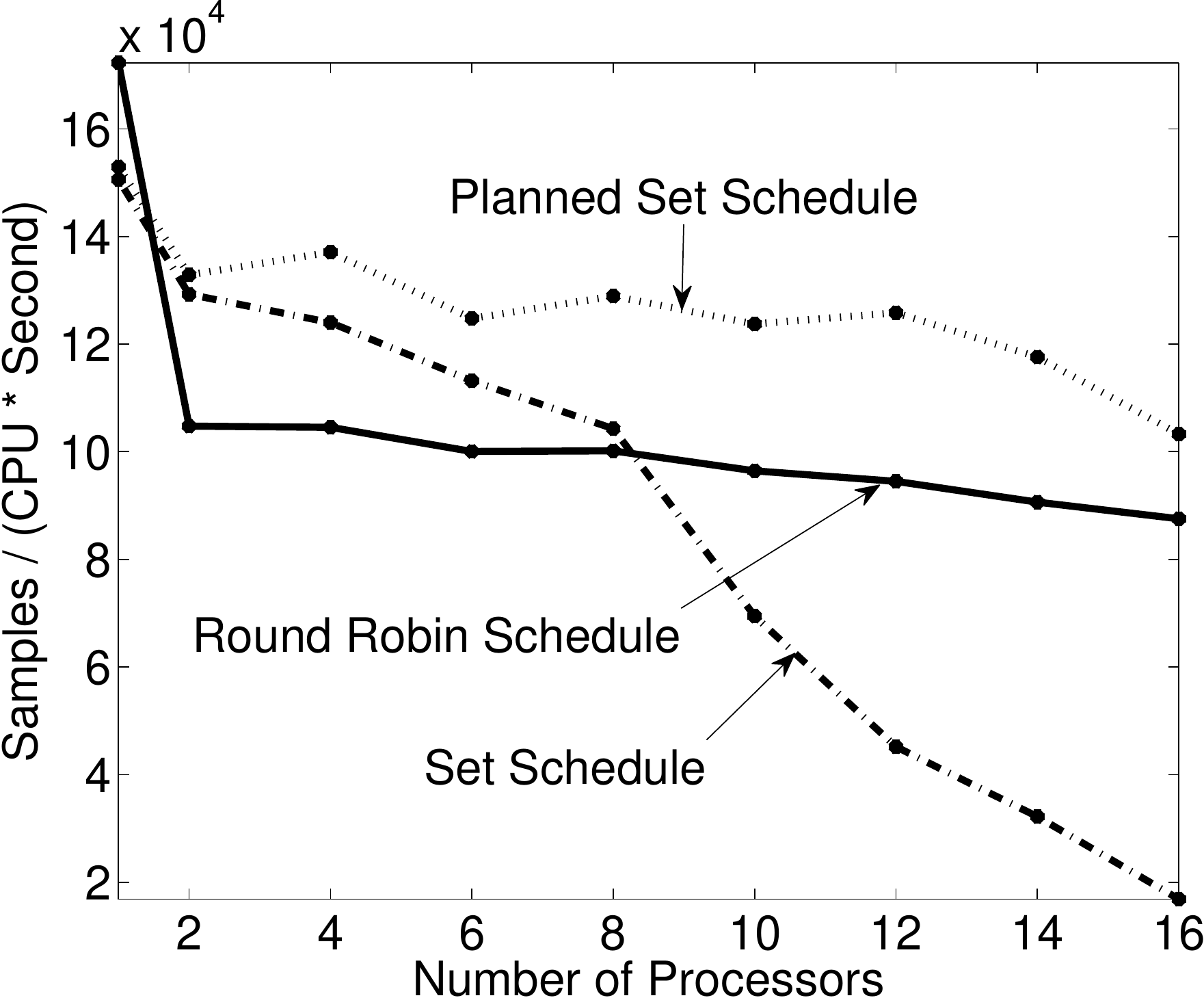}
     }
     \subfigure[BP Speedup] { 
       \label{fig:bpspeedup}
       \includegraphics[width=.185\textwidth]{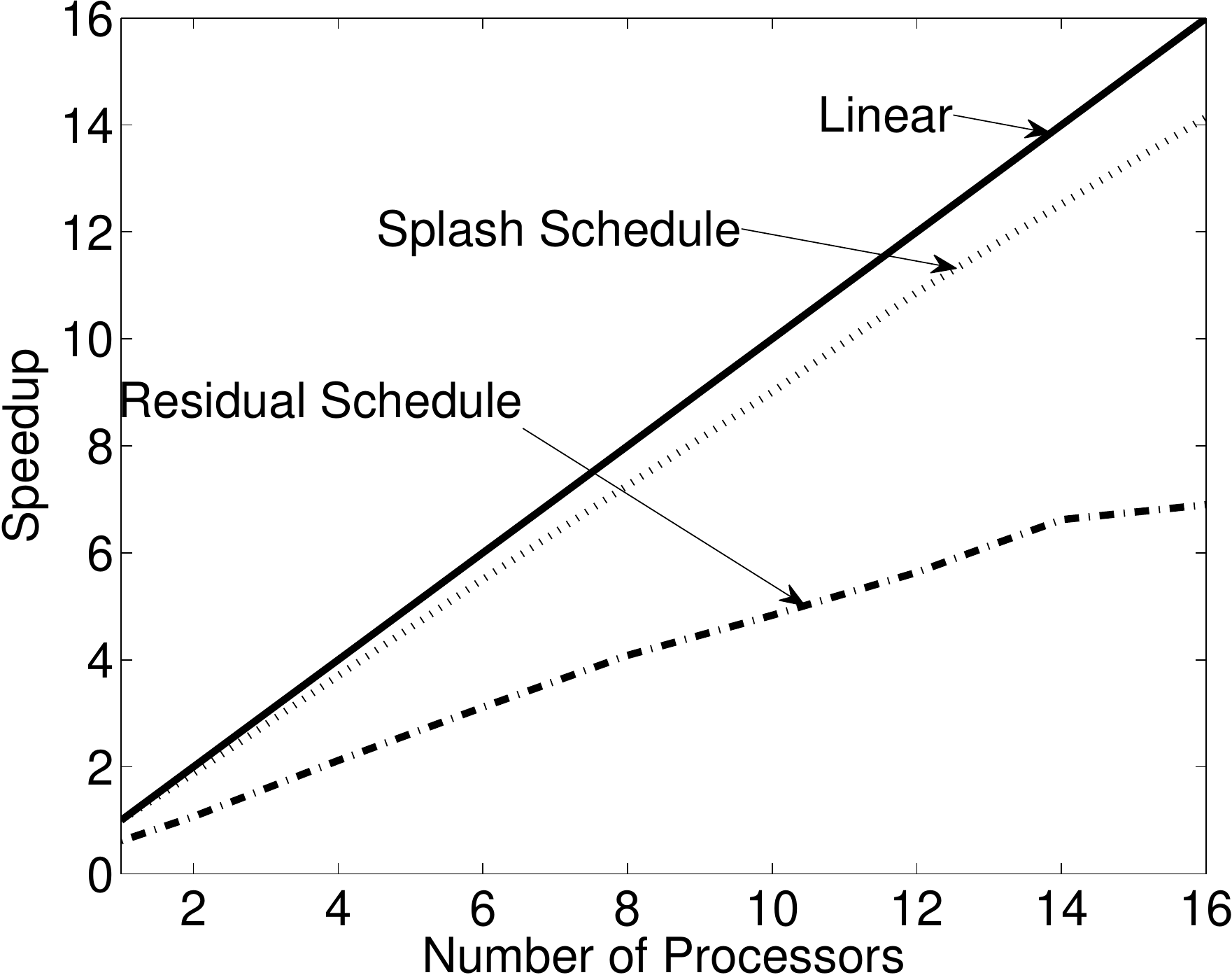}
     }
     \subfigure[BP Eff.] { 
       \label{fig:bpeff}
       \includegraphics[width=.185\textwidth]{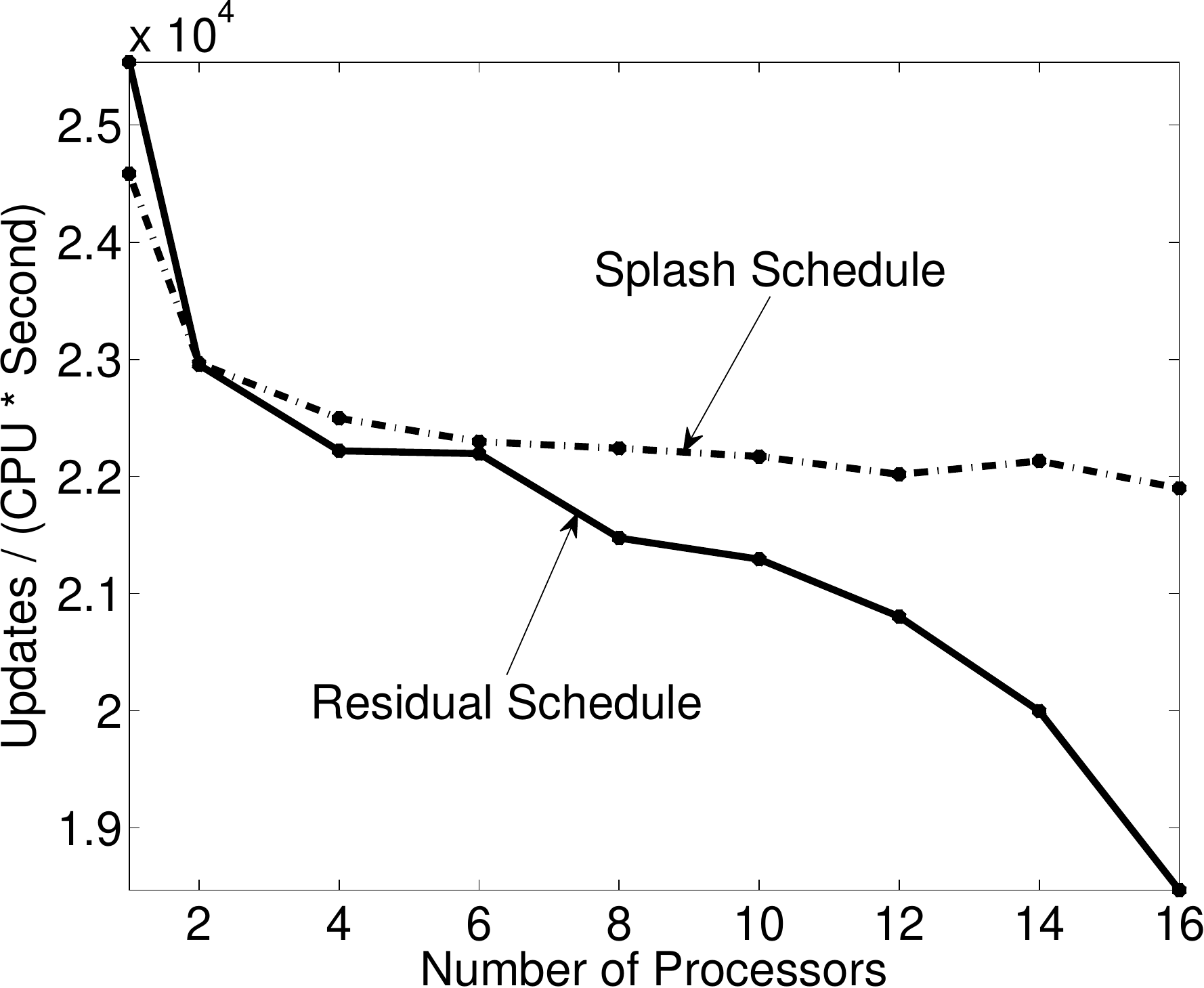}
     }
     \caption{ \footnotesize \emph{MRF Inference} \textbf{(a)} The
       speedup of the Gibbs sampler using three different schedules.
       The \emph{planned set schedule} enables processors to safely
       execute more than one color simultaneously.  The \emph{round
         robin schedule} executes updates in a fixed order and relies
       on the edge consistency model to maintain sequential
       consistency.  The plan \emph{set scheduler} does not apply
       optimization and therefore suffers from substantial
       synchronization overhead.  \textbf{(b)} The distribution of
       vertices over the 20 colors is strongly skewed resulting in a
       high sequential set schedule.  \textbf{(c)} The sampling rate
       per processor plotted against the number of processor provides
       measure of parallel overhead which is substantially reduced by
       the plan optimization in the set scheduler. \textbf{(d)} The
       speedup for Loopy BP is improved substantially by the Splash.
       \textbf{(e)} The efficiency of the GraphLab framework as
       function of the number of processors.  }
     \label{fig:gibbsbp}
   \end{center}
\vspace{-0.5cm}
\end{figure*}

\paragraph{Results:} 
In \figref{fig:gibbsbp} we present the speedup and efficiency results
for Gibbs sampling and Loopy BP.  Using the set schedule in
conjunction with the planning optimization enables the Gibbs sampler
to achieve a factor of 10 speedup on 16 processors.  The execution
plan takes 0.05 seconds to compute, an immaterial fraction of the 16
processor running time.  Because of the structure of the MRF, a large
number of colors (20) is needed and the vertex distribution over
colors is heavily skewed.  Consequently there is a strong sequential
component to running the Gibbs sampler on this model.  In contrast the
Loopy BP speedup demonstrates considerably better scaling with factor
of 15 speedup on 16 processor.  The larger cost per BP update in
conjunction with the ability to run a fully asynchronous schedule
enables Loopy BP to achieve relatively uniform update efficiency
compared to Gibbs sampling.


\tightsubsection{Co-EM}

To illustrate how GraphLab scales in settings with large structured
models we designed and implemented a parallel version of Co-EM
\citep{CoEMJones, CoEMGhani}, a semi-supervised learning algorithm for
named entity recognition (NER).  Given a list of noun phrases (NP)
(e.g., ``big apple''), contexts (CT) (e.g., ``citizen of
\underline{\hspace*{2mm}}''), and co-occurence counts for each NP-CT
pair in a training corpus, CoEM tries to estimate the probability
(belief) that each entity (NP or CT) belongs to a particular class
(e.g., ``country'' or ``person'').  The CoEM update function is
relatively fast, requiring only a few floating operations, and
therefore stresses the GraphLab implementation by requiring GraphLab
to manage massive amounts of fine-grained parallelism.


The GraphLab graph for the CoEM algorithm is a bipartite graph with
each NP and CT represented as a vertex, connected by edges with
weights corresponding to the co-occurence counts. Each vertex stores
the current estimate of the belief for the corresponding entity.  The
update function for CoEM recomputes the local belief by taking a
weighted average of the adjacent vertex beliefs. The adjacent vertices
are rescheduled if the belief changes by more than some predefined
threshold ($10^{-5}$).


We experimented with the following two NER datasets obtained from
web-crawling data.


\begin{center}
  \footnotesize
  \begin{tabular}{|l|l|l|l|l|}
    \hline
    Name & Classes & Verts. & Edges & 1 CPU Runtime \\
    \hline 
    small & 1 &  0.2 mil. & 20 mil. & 40 min  \\
    large & 135 &  2 mil. & 200 mil. & 8 hours \\
    \hline
  \end{tabular}
\end{center}

\begin{figure*}[t]
  \begin{center}
    \subfigure[CoEM Speedup Small]{ 
      \label{fig:CoemScalabilityPlota} 
      \includegraphics[width=.23\textwidth]{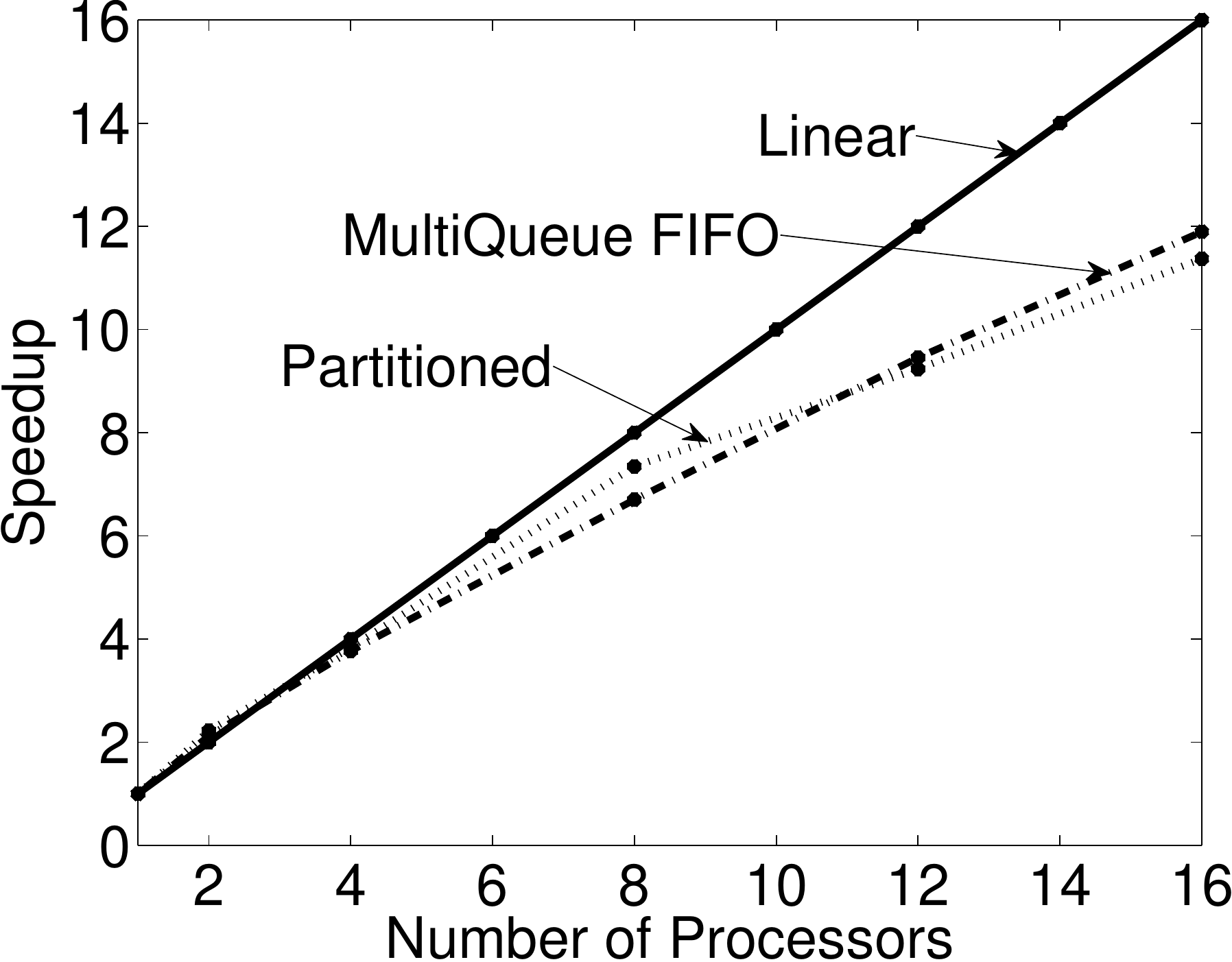}
    }
    \subfigure[CoEM Speedup Large]{ 
      \label{fig:CoemScalabilityPlotb} 
      \includegraphics[width=.23\textwidth]{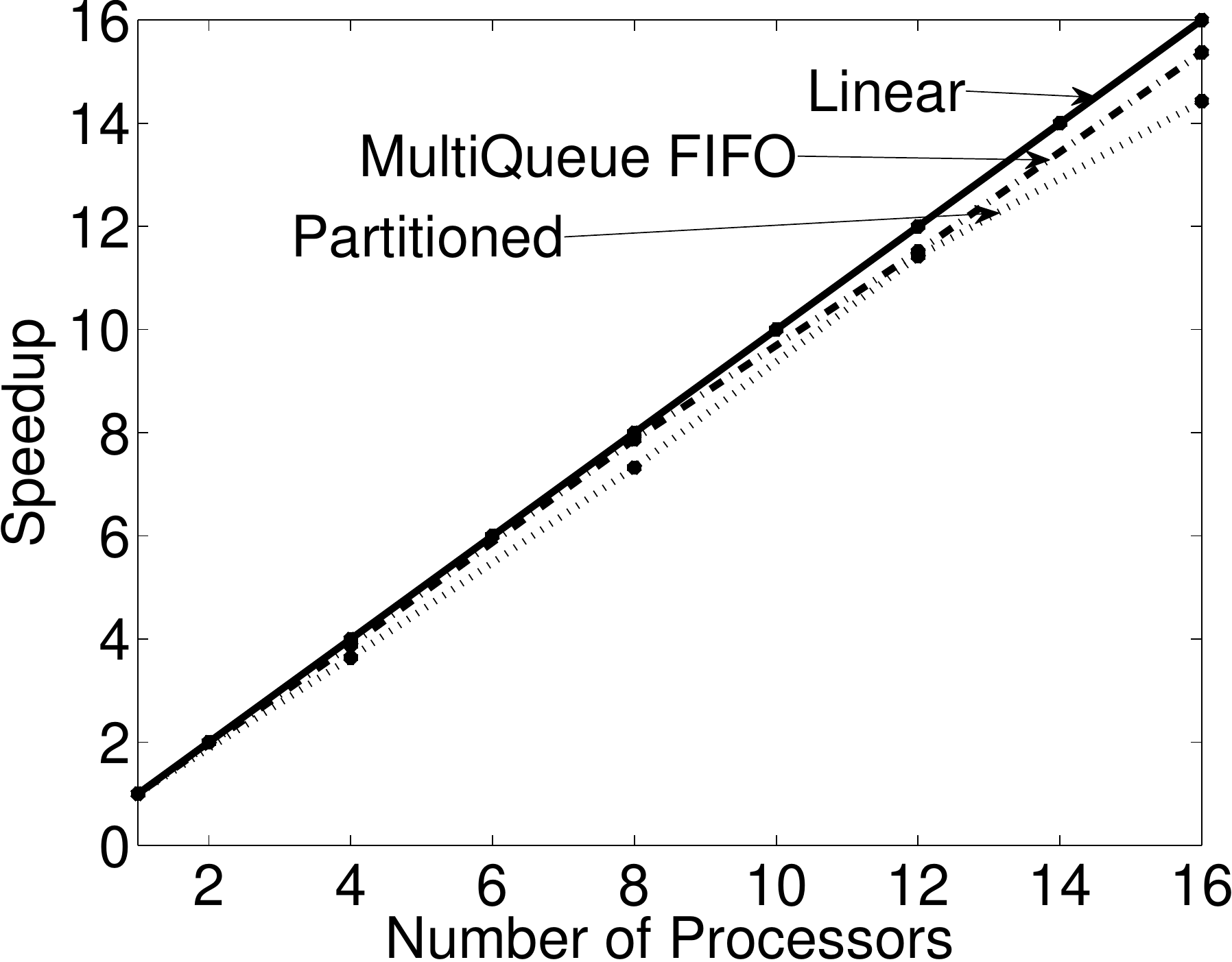}      
    }
    \subfigure[Convergence]{
      \includegraphics[width=.23\textwidth]{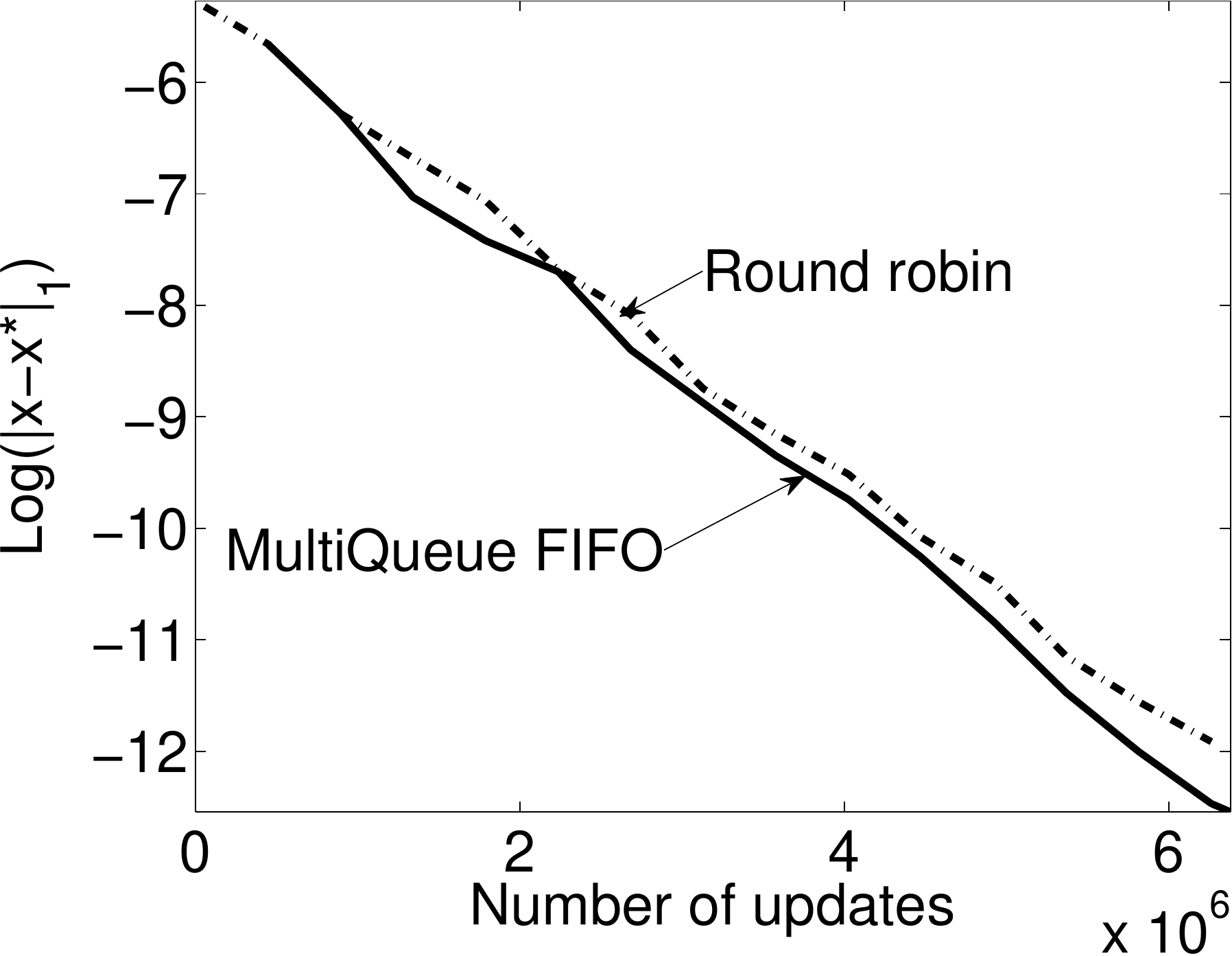}
      \label{fig:CoEMConvergence} }
    \subfigure[Speedup with Problem Size]{
      \includegraphics[width=.23\textwidth]{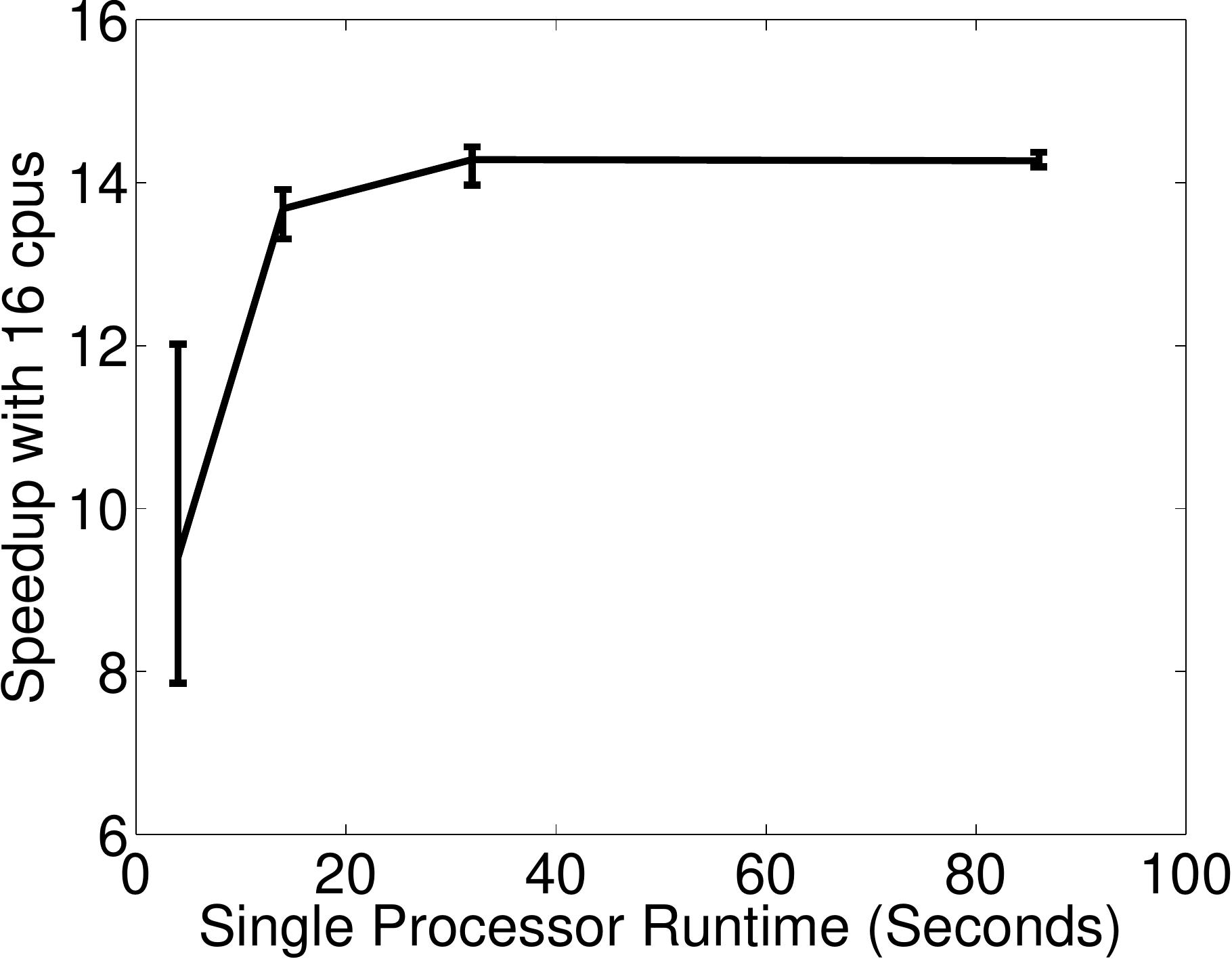} 
       \label{fig:CoEMProbSize} }

  \end{center}
  \caption{ \footnotesize \emph{CoEM Results} \textbf{(a,b)} Speedup
    of MultiQueue FIFO and Partitioned Scheduler on both
    datasets. Speedup is measured relative to fastest running time on
    a single cpu. The large dataset achieves better scaling because
    the update function is slower. \textbf{(c)} Speed of convergence
    measured in number of updates for MultiQueue FIFO and Round Robin
    (equivalent to synchronized Jacobi schedule), \textbf{(d)} Speedup
    achieved with 16 cpus as the graph size is varied. }
\end{figure*}

We plot in \figref{fig:CoemScalabilityPlota} and
\figref{fig:CoemScalabilityPlotb} the speedup obtained by the
Partitioned Scheduler and the MultiQueue FIFO scheduler, on both small
and large datasets respectively.  We observe that both schedulers
perform similarly and achieve nearly linear scaling. In addition, both
schedulers obtain similar belief estimates suggesting that the update
schedule may not affect convergence in this application.


With 16 parallel processors, we could complete three full Round-robin
iterations on the large dataset in less than 30 minutes. As a
comparison, a comparable Hadoop implementation took approximately 7.5
hours to complete the exact same task, executing on an average of 95
cpus.  [Personal communication with Justin Betteridge and Tom
Mitchell, Mar 12, 2010].  Our large performance gain can be attributed
to data persistence in the GraphLab framework.  Data persistence
allows us to avoid the extensive data copying and synchronization
required by the Hadoop implementation of MapReduce.

Using the flexibility of the GraphLab framework we were able to study
the benefits of dynamic (Multiqueue FIFO) scheduling versus a regular
round-robin scheduling in CoEM. \figref{fig:CoEMConvergence}
compares the number of updates required by both schedules to obtain a
result of comparable quality on the larger dataset. Here we measure
quality by $\Lone$ parameter distance to an empirical estimate of the
fixed point $x^*$, obtained by running a large number of synchronous
iterations.  For this application we do not find a substantial benefit
from dynamic scheduling.




We also investigated how GraphLab scales with problem size. Figure
\ref{fig:CoEMProbSize} shows the maximum speedup on 16 cpus attained
with varying graph sizes, generated by sub-sampling a fraction of
vertices from the large dataset.  We find that parallel scaling
improves with problem size and that even on smaller problems GraphLab
is still able to achieve a factor of 12 speedup on 16 cores.


\tightsubsection{Lasso}

The Lasso \citep{tibshirani96regression} is a popular feature
selection and shrinkage method for linear regression which minimizes
the objective $L(w) = \sum_{j=1}^n (w^T x_j - y_j)^2 + \lambda
\LOneNorm{w}$.  Unfortunately, there does not exist, to the best of
our knowledge, a parallel algorithm for fitting a Lasso model. In this
section we implement 2 different parallel algorithms for solving the
Lasso.

\subsubsection{Shooting Algorithm}
We use GraphLab to implement the Shooting Algorithm \citep{Shooting},
a popular Lasso solver, and demonstrate that GraphLab is able to
\emph{automatically} obtain parallelism by identifying operations that
can execute concurrently while retaining sequential consistency.

The shooting algorithm works by iteratively minimizing the objective
with respect to each dimension in $w$, corresponding to coordinate
descent.  We can formulate the Shooting Algorithm in the GraphLab
framework as a bipartite graph with a vertex for each weight $w_i$ and
a vertex for each observation $y_j$. An edge is created between $w_i$
and $y_j$ with weight $X_{i,j}$ if and only if $X_{i,j}$ is non-zero.
We also define an update function (\algref{alg:shooting}) which
operates only on the weight vertices, and corresponds exactly to a
single minimization step in the shooting algorithm. A round-robin
scheduling of \algref{alg:shooting} on all weight vertices corresponds
exactly to the sequential shooting algorithm.  We automatically obtain
an equivalent parallel algorithm by select the full consistency model.
Hence, by encoding the shooting algorithm in GraphLab we are able to
discover a sequentially consistent \emph{automatic parallelization}.

We evaluate the performance of the GraphLab implementation on a
financial data set obtained from \cite{shimon09}. The task is to use word
counts of a financial report to predict stock volatility of the
issuing company for the consequent 12 months.  Data set consists of
word counts for 30K reports with the related stock volatility metrics.

\begin{figure}
   \begin{center}
     \subfigure[Sparser Dataset Speedup] {
       \label{fig:100speedup}
       \includegraphics[width=.22\textwidth]{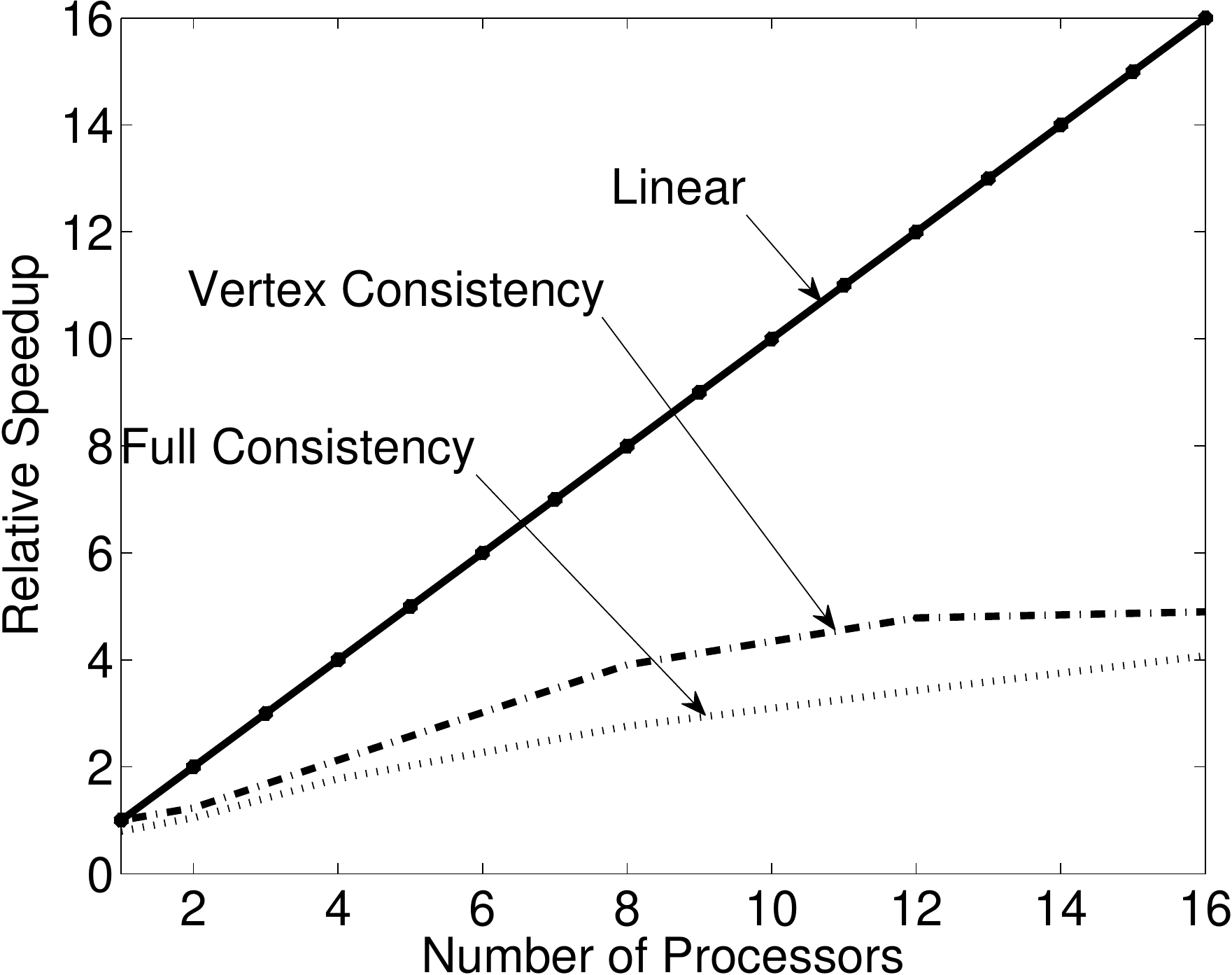}
     }
     \subfigure[Denser Dataset Speedup] {
       \label{fig:1000speedup}
       \includegraphics[width=.22\textwidth]{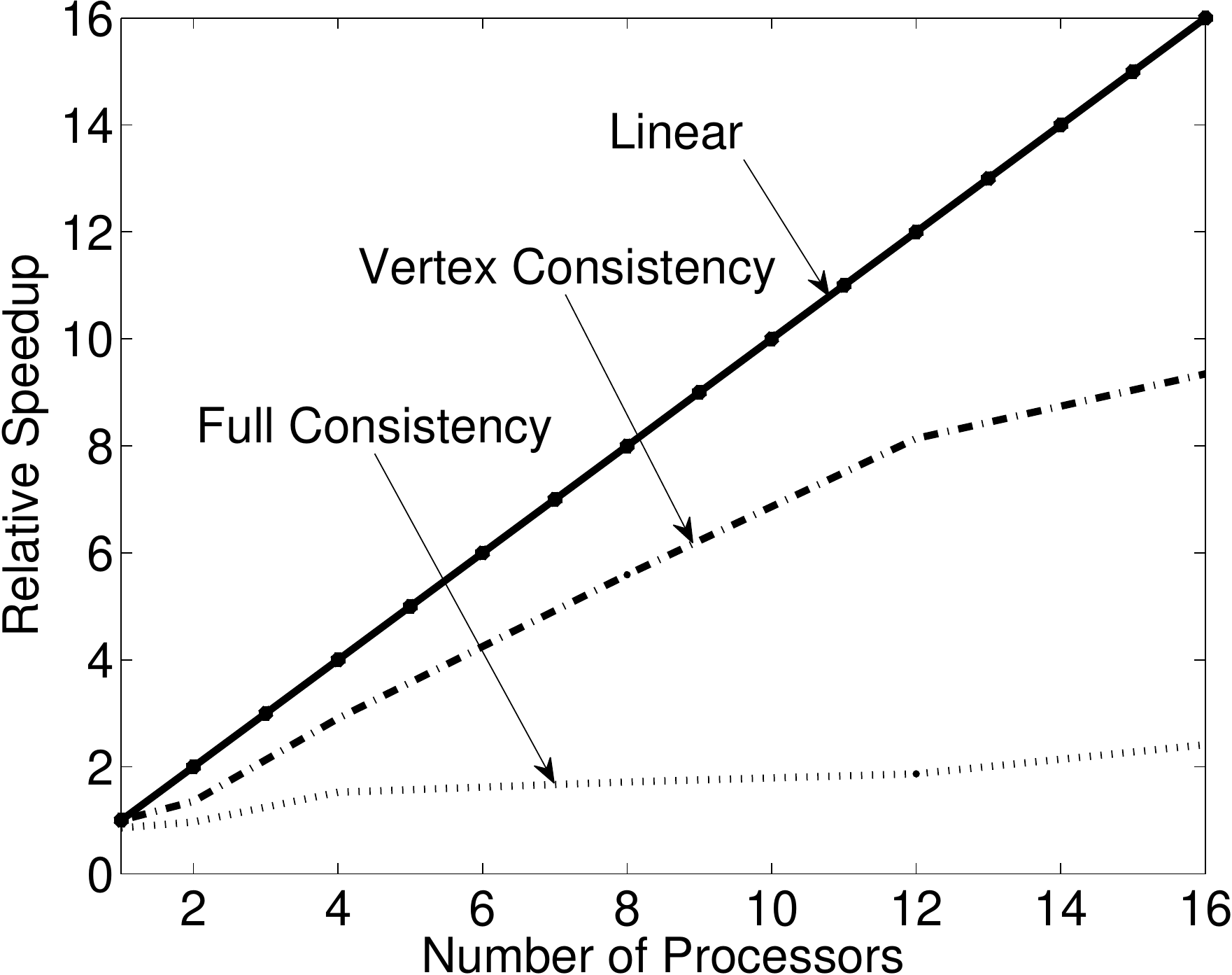}
     }
     \caption{ \footnotesize \emph{Shooting Algorithm} \textbf{(a)}
       Speedup on the sparser dataset using \textit{vertex
         consistency} and \textit{full consistency} relative to the
       fastest single processor runtime. \textbf{(b)} Speedup on the
       denser dataset using \textit{vertex consistency} and
       \textit{full consistency} relative to the fastest single
       processor runtime. }
     \label{fig:lasso}
   \end{center}
   \vspace{-5mm}
\end{figure}

To demonstrate the scaling properties of the full consistency model,
we create two datasets by deleting common words. The sparser dataset
contains 209K features and 1.2M non-zero entries, and the denser
dataset contains 217K features and 3.5M non-zero entries.  The speedup
curves are plotted in \figref{fig:lasso}. We observed better scaling
(4x at 16 CPUs) on the sparser dataset than on the denser dataset (2x
at 16 CPUs).  This demonstrates that ensuring full consistency on
denser graphs inevitably increases contention resulting in reduced
performance.

Additionally, we experimented with relaxing the consistency model, and
we discovered that the shooting algorithm still converges under the
weakest vertex consistency guarantees; obtaining solutions with only
$0.5\%$ higher loss on the same termination criterion.  The vertex
consistent model is much more parallel and we can achieve
significantly better speedup, especially on the denser dataset. It
remains an open question why the Shooting algorithm still functions
under such weak guarantees.

\begin{algorithm}[t]
  \footnotesize
  \caption{Shooting Algorithm}
  \dontprintsemicolon
  \SetLine
  ShootingUpdate($\vertexdata{w_i}, \vertexdatain{w_i},\vertexdataout{w_i}$)
  \Begin{
    Minimize the loss function with respect to $w_i$ \;
    \If{$w_i$ changed by $>\epsilon$}{
      Revise the residuals on all $y's$ connected to $w_i$ \;
      Schedule all $w's$ connected to neighboring $y's$
    }
  }
  \label{alg:shooting}
\end{algorithm}

\subsection{Compressed Sensing}
To show how GraphLab can be used as a subcomponent of a larger
\emph{sequential} algorithm, we implement a variation of the
interior point algorithm proposed by \cite{L1Boyd} for the purposes
of compressed sensing. The aim is to use a sparse linear
combination of basis functions to represent the image, 
while minimizing the reconstruction error. Sparsity is 
achieved through the use of elastic net regularization.

The interior point method is a double loop algorithm where the
sequential outer loop (\algref{alg:outerloop}) implements a Newton
method while the inner loop computes the Newton step by solving a
sparse linear system using GraphLab.  We used Gaussian BP (GaBP) as a
linear solver \citep{bicksonThesis} since it has a natural GraphLab
representation.  The GaBP GraphLab construction follows closely the BP
example in \secref{sec:eyeball}, but represents potentials and
messages analytically as Gaussian distributions.  In addition, the
outer loop uses a Sync operation on the data graph to compute the
duality gap and to terminate the algorithm when the gap falls below a
predefined threshold.  Because the graph structure is fixed across
iterations, we can leverage data persistency in GraphLab, avoid both
costly set up and tear down operations and resume from the converged
state of the previous iteration.

\begin{figure}[t]
   \begin{center}
     \subfigure[Speedup] {
       \label{fig:l1speedup}
       \includegraphics[width=.20\textwidth]{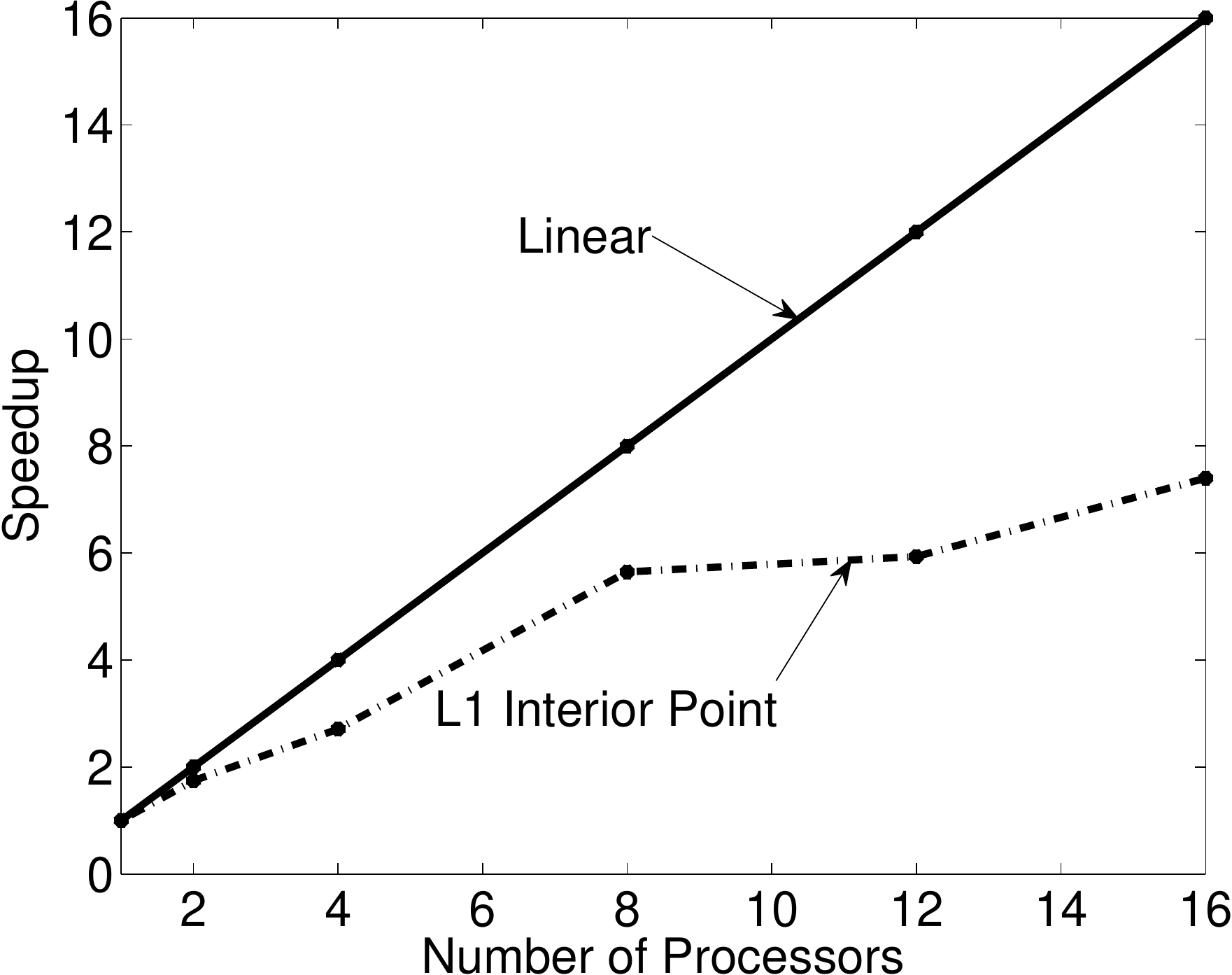}
     }\hspace{-1mm}
     \subfigure[Lenna] {
       \label{fig:gibbscolor}
       \includegraphics[width=2.1cm, height=2.6cm]{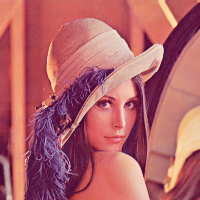}
     }\hspace{-2mm}
     \subfigure[Lenna 50\%] {
       \label{fig:gibbseff}
       \includegraphics[width=2.1cm, height=2.6cm]{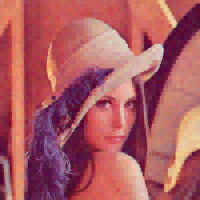}
     }
     \caption{ \footnotesize \textbf{(a)} Speedup of the Interior Point
     algorithm on the compressed sensing dataset,  \textbf{(b)}
     Original 256x256 test image with 65,536 pixels,
     \textbf{(c)} Output of compressed sensing algorithm using 32,768
     random projections.}
     \label{fig:l1results}
   \end{center}
   \vspace{-4mm}
\end{figure}

We evaluate the performance of this algorithm on a synthetic
compressed sensing dataset constructed by applying a random projection
matrix to a wavelet transform of a $256 \times 256$ Lenna image
(\figref{fig:l1results}). Experimentally, we achieved a factor of 8
speedup using 16 processors using the round-robin scheduling.



\begin{algorithm}[t]
  \footnotesize
  \caption{Compressed Sensing Outer Loop}
  \dontprintsemicolon
  \SetLine  
   \While{$duality\_gap \ge \epsilon$}{
     Update edge and node data of the data graph.  \;
     Use GraphLab to run GaBP on the graph \;
     Use Sync to compute duality gap \;
     Take a newton step \;
   }
  \label{alg:outerloop}
\end{algorithm}

%


\tightsection{Conclusions and Future Work}

We identified several limitations in applying existing parallel
abstractions like MapReduce to Machine Learning (ML) problems. By
targeting common patterns in ML, we developed \term{GraphLab}, a new
parallel abstraction which achieves a high level of usability,
expressiveness and performance.  Unlike existing parallel
abstractions, GraphLab supports the representation of structured data
dependencies, iterative computation, and flexible scheduling.

The \term{GraphLab} abstraction uses a \term{data graph} to encode the
computational structure and data dependencies of the problem.
GraphLab represents local computation in the form of \term{update
  functions} which transform the data on the \term{data
  graph}. Because update functions can modify overlapping state, the
GraphLab framework provides a set of data \term{consistency models}
which enable the user to specify the minimal consistency requirements
of their application without having to build their own complex locking
protocols.  To manage sharing and aggregation of global state,
GraphLab provides a powerful \term{sync mechanism}.

To manage the scheduling of dynamic iterative parallel computation,
the GraphLab abstraction provides a rich collection of parallel
\term{schedulers} encompassing a wide range of ML algorithms.
GraphLab also provides a scheduler construction framework built around
a sequence of vertex sets which can be used to compose custom
schedules.

We developed an optimized shared memory implementation GraphLab and we
demonstrated its performance and flexibility through a series of case
studies.  In each case study we designed and implemented a popular ML
algorithm and applied it to a large real-world dataset achieving
state-of-the-art performance.

Our ongoing research includes extending the GraphLab framework to the
distributed setting allowing for computation on even larger datasets.
While we believe GraphLab naturally extend to the distributed setting
we face numerous new challenges including efficient graph
partitioning, load balancing, distributed locking, and fault
tolerance.

\subsubsection*{Acknowledgements}
\vspace{-1mm} We thank Guy Blelloch and David O'Hallaron for their
guidance designing and implementing GraphLab. This work is supported
by ONR Young Investigator Program grant N00014-08-1-0752, the ARO
under MURI W911NF0810242, DARPA IPTO FA8750-09-1-0141, and the NSF
under grants IIS-0803333 and NeTS-NBD CNS-0721591.  Joseph Gonzalez is
supported by the AT\&T Labs Fellowship Program.

\vspace{-3mm}
{\denselistbib
  \footnotesize
  \bibliographystyle{unsrtnat}
  \bibliography{references}
}

\end{document}